\pdfoutput=1
\documentclass{article} % For LaTeX2e
\usepackage[table,xcdraw,dvipsnames]{xcolor}
\usepackage{iclr2023_conference,times}
\usepackage{amsmath,amsfonts,bm}

\def\eqref#1{equation~\ref{#1}}
\def\1{\bm{1}}

\DeclareMathAlphabet{\mathsfit}{\encodingdefault}{\sfdefault}{m}{sl}
\SetMathAlphabet{\mathsfit}{bold}{\encodingdefault}{\sfdefault}{bx}{n}

\usepackage{hyperref}
\usepackage{url}

\usepackage{enumitem} 
\usepackage{tabularx}
\usepackage{booktabs}
\usepackage{multirow}

\usepackage{graphicx}
\usepackage{balance}
\usepackage{adjustbox}
\usepackage{mathtools}
\usepackage{color}

\usepackage{epsfig}
\usepackage{amsmath}
\usepackage{amssymb}
\usepackage{algorithm,algpseudocode}
\usepackage{tabularx}
\usepackage{booktabs}
\usepackage{makecell}
\usepackage{tabularx}
\usepackage{listings}

\newcommand{\ie}{\textit{i}.\textit{e}.}
\newcommand{\eg}{\textit{e}.\textit{g}.}

\newcommand{\tablestyle}[2]{\setlength{\tabcolsep}{#1}\renewcommand{\arraystretch}{#2}\centering\small}
\newcommand{\highlight}[1]{\cellcolor{Gray!16}\textbf{#1}}
\newcommand{\gain}[1]{\footnotesize\textcolor{Green}{(+{#1})}}% \textbf

\definecolor{mydarkblue}{rgb}{0,0.1,0.6}
\hypersetup{
    colorlinks=true,
    citecolor=mydarkblue
          }

\title{Is synthetic data from generative models ready for image recognition?}

\vspace{-0.4cm}
\author{
Ruifei He\textsuperscript{\rm 1*} 
\;\; Shuyang Sun\textsuperscript{\rm 2} 
\;\; Xin Yu\textsuperscript{\rm 1} 
\;\; Chuhui Xue\textsuperscript{\rm 3} 
\;\; Wenqing Zhang\textsuperscript{\rm 3} 
\;\; Philip Torr\textsuperscript{\rm 2} \\
\;\textbf{Song Bai}\textsuperscript{\rm 3$\dagger$}
\;\; \textbf{Xiaojuan Qi}\textsuperscript{\rm 1$\dagger$}
\\
\;\textsuperscript{\rm 1}The University of Hong Kong \ \textsuperscript{\rm 2}University of Oxford \ \textsuperscript{\rm 3}ByteDance 
}

\iclrfinalcopy % Uncomment for camera-ready version, but NOT for submission.
\begin{document}

\maketitle

{\let\thefootnote\relax\footnotetext{$^*$ Part of the work is done during an internship at ByteDance. Email: \href{mailto:ruifeihe@eee.hku.hk}{\color{black}{ruifeihe@eee.hku.hk}}}}
{\let\thefootnote\relax\footnotetext{$^\dagger$ Corresponding authors: 
\href{mailto:songbai.site@gmail.com}{\color{black}{songbai.site@gmail.com}}, \href{mailto:xjqi@eee.hku.hk}{\color{black}{xjqi@eee.hku.hk}}}}

\vspace{-0.6cm}
\begin{abstract}
Recent text-to-image generation models have shown promising results in generating high-fidelity photo-realistic images. Though the results are astonishing to human eyes, how applicable these generated images are for recognition tasks remains under-explored. In this work, we extensively study whether and how synthetic images generated from state-of-the-art text-to-image generation models can be used for image recognition tasks, and focus on two perspectives: synthetic data for improving classification models in data-scarce settings ({\ie} zero-shot and few-shot), and synthetic data for large-scale model pre-training for transfer learning. We showcase the powerfulness and shortcomings of synthetic data from existing generative models, and propose strategies for better applying synthetic data for recognition tasks. Code:  \href{https://github.com/CVMI-Lab/SyntheticData}{\color{blue}{https://github.com/CVMI-Lab/SyntheticData}.}
\end{abstract}

\vspace{-0.4cm}
\section{Introduction}

Over the past decade, deep learning powered by large-scale annotated data has revolutionized the field of image recognition. 
However, it is costly and time-consuming to manually collect a large-scale labeled dataset, and recent concerns about data privacy and usage rights further hinder this process. 
In parallel, generative models that aim to model real-data distributions can now produce high-fidelity photo-realistic images. In particular, recent text-to-image generation models \citep{nichol2021glide, ramesh2022hierarchical, saharia2022photorealistic} have made major breakthroughs in synthesizing high-quality images from text descriptions. 
This promotes us to ask: is synthetic data from generative models ready for image recognition tasks? 

There are a few early attempts at exploring synthetic data from generative models for image recognition tasks. \citet{besnier2020dataset} use a class-conditional GAN (BigGAN \citep{brock2018large} trained for ImageNet-1000 classes) to generate images for training image classifiers. 
\citet{zhang2021datasetgan} leverage StyleGAN \citep{karras2019style} to produce synthetic labeled data for object-part segmentation. 
\citet{jahanian2021generative} manipulate the latent space of a GAN model to produce multi-view images for contrastive learning. 
Albeit promising, early works either address tasks on a small scale or only for a specific setting. Plus, they all focus on GAN-based models and none explore the revolutionary text-to-image generation models, which hold more promises to benefit recognition tasks.

\renewcommand\thefootnote{\textcolor{red}{\arabic{footnote}}}

In this paper, we present the first study on the state-of-the-art text-to-image generation models for image recognition.
With the power of text-to-image generation, we could hopefully not only generate massive high-quality labeled data, but also achieve domain customization by generating synthetic data targeted for a specific label space, {\ie} the label space of a downstream task. 
Our study is carried out on one open-sourced text-to-image generation model, GLIDE \citep{nichol2021glide} \footnote{At the beginning of this project, GLIDE
% (\url{https://github.com/openai/glide-text2im}) 
is the only open-sourced text-to-image synthesis model that also delivers high-quality synthesis results.}. 
We attempt to uncover the benefits and pitfalls of synthetic data for image recognition through the lens of investigating the following two questions: 1) is synthetic data from generative models ready for improving classification models? 2) %is synthetic data from generative models ready for  large-scale model pre-training for transfer learning? 
whether synthetic data can be a feasible source for transfer learning ({\ie} model pre-training)? 
It is worth noting that for 1), we only studied the zero-shot and few-shot settings because the positive impact of synthetic data diminishes as more shots are present. And, we build most of our investigations on the state-of-the-art method CLIP \citep{radford2021learning} with the feature extractor initialized with large-scale pre-trained weights frozen. % reason is that naively 

\noindent \textbf{Our Findings.}
 First, in the zero-shot setting, {\ie} no real-world data are available, we demonstrate that synthetic data can significantly improve classification results on $17$ diverse datasets: the performance is increased by $4.31\%$ in top-1 accuracy on average, and even improved by as much as $17.86\%$ on the EuroSAT dataset. 
To better leverage synthetic data in this setting, we also investigate useful strategies to increase data diversity, reduce data noise, and enhance data reliability. This is achieved by designing diversified text prompts and measuring the correlation of text and synthesized data with CLIP features. 

Second, in the few-shot setting, {\ie} a few real images are available, albeit not as significant as in the zero-shot task, synthetic data are also shown to be beneficial and help us achieve a new state of the art.
Our observation shows that the domain gap between synthetic data and downstream task data is one challenge on further improving the effectiveness of synthetic data on classifier learning.
Fortunately, in this setting, the accessibility of real data samples can provide useful information about the data distribution of the downstream task. We thus propose to use real images as guidance in the generation process to reduce domain gaps and improve effectiveness.

Third, in large-scale model pre-training for transfer learning, our study shows that synthetic data are suitable and effective for model pre-training, delivering superior transfer learning performance and even outperforming ImageNet pre-training. Especially, synthetic data work surprisingly well in unsupervised model pre-training, and favor ViT-based backbones. We also demonstrate that by increasing the label space ({\ie} text prompts) for data generation, the enlarged data amount and diversity could further bring performance boosts.
Besides, synthetic data can work collaboratively with real data ({\ie} ImageNet) where we obtain improved performance when the model is initialized with ImageNet pre-trained weights.

\vspace{-0.2cm}
\section{Related works}
\vspace{-0.1cm}

\noindent \textbf{Synthetic Data for Image Recognition.}
There are mainly two forms of synthetic data for image recognition, {\ie} 1) synthetic datasets generated from a traditional simulation pipeline; 2) synthetic images output from generative models. 

The first type, synthetic datasets \citep{DFIB15, peng2017visda, richter2016playing}, are usually generated from a traditional pipeline with a specific data source, 
 \eg synthetic 2D renderings of 3D models or scenes from graphics engines. 
However, this traditional way of generating synthetic datasets has several drawbacks: 1) manually defined pipeline generated synthetic data may have a  certain gap with real-world data; 2) taking up huge physical space to store and huge cost to share and transfer; 3) data amount and diversity bounded by the specific data source. 

Compared with synthetic datasets, generative models are a more efficient means of synthetic data representation, exhibiting favorable advantages: 1) could produce high-fidelity photorealistic images closer to real data since they are trained on real-world data; 2) highly condensed compared to synthetic data itself, and take up much reduced storage space; 3) potentially unlimited synthetic data size. 
Only recently, few works attempt to explore synthetic data generated from generative models for image recognition. \citet{besnier2020dataset} use a class-conditional GAN to train classifiers of the same classes. 
\citet{zhang2021datasetgan} leverage the latent code of StyleGAN \citep{karras2019style} to produce labels for object part segmentation. While they achieve promising results, both works are task-wise and only employed on a small scale.
\citet{jahanian2021generative} use a GAN-based generator to generate multiple views to conduct unsupervised contrastive representation learning. These works, however, explore upon the traditional GAN-based models; in contrast, our work investigates with the best released text-to-image generation model, which demonstrates new customization ability for different downstream label space.

\noindent \textbf{Text-to-Image Diffusion Models.} \label{sec: text-diffusion}
Diffusion models \citep{sohl2015deep, ho2020denoising, nichol2021improved} have recently emerged as a class of promising and powerful generative models. 
As a likelihood-based model, the diffusion model matches the underlying data distribution $q(x_0)$ by learning to reverse a noising process, and thus novel images can be sampled from a prior Gaussian distribution via the learned reverse path. Because of the high sample quality, good mode coverage and promising training stability, diffusion models are quickly becoming a new trend in both unconditional \citep{ho2020denoising, nichol2021improved, ho2022cascaded} and conditional \citep{dhariwal2021diffusion, rombach2022high, lugmayr2022repaint, saharia2022palette, meng2021sdedit, saharia2022image} image synthesis fields.

In particular, text-to-image generation can be treated as a conditional image generation task that requires the sampled image to match the given natural language description. Based upon the formulation of the diffusion model, several text-to-image models such as Stable diffusion \citep{rombach2022high}, DALL-E2 \citep{ramesh2022hierarchical}, Imagen \citep{saharia2022photorealistic} and GLIDE \citep{nichol2021glide} deliver unprecedented synthesis quality, largely facilitating the development of the AI-for-Art community. Despite achieving astonishing perceptual results, their potential utilization for high-level tasks is yet under-explored. In this paper, we utilize the state-of-the-art model GLIDE and showcase its powerfulness and shortcomings for synthesizing data for recognition tasks.

\vspace{-0.2cm}
\section{Is synthetic data ready for image recognition?} \label{sec:3}
\vspace{-0.15cm}

In the following sections, we answer the question by studying whether synthetic data can benefit recognition tasks and how to better leverage synthetic data to address different tasks. 
We carry out our exploration through the lens of two basic settings with three tasks: synthetic data for improving classification models in the data-scarce setting ({\ie} zero-shot and few-shot) (see Sec. \ref{sec: zero-shot} and Sec. \ref{sec: few-shot}) and synthetic data for model pre-training for transfer learning (see Sec. \ref{sec: pre-train}).

\textbf{Model Setup for Data-scarce ({{\ie} Zero-shot and Few-shot}) Image Classification.}
As CLIP \citep{radford2021learning} is the state-of-the-art approach for zero-shot learning, we conduct our study for zero-shot and few-shot settings upon pre-trained CLIP models, aiming to better understand synthetic data upon strong baselines. There have been a few attempts on better tuning pre-trained CLIP for data-scarce image classification, such as CoOp \citep{zhou2022learning}, CLIP Adapter \citep{gao2021clip}, and Tip Adapter \citep{zhang2022tip}, where the image encoder is frozen for better preserving the pre-trained feature space.
We argue that different tuning methods could all be regarded as different ways of learning classifier weights, {\eg} CoOp optimizes learnable prompts for better learning classifiers. 

Here, we adopt a simple tuning method, Classifier Tuning (\textbf{CT}), a baseline method introduced in \citet{wortsman2022robust}.
Concretely, for a k-way classification, we input the class names $C=\{c_1,...,c_k\}$ with prompt $s_i=``\text{a photo of a }\{c_i\}"$ into the text encoder $h$ of CLIP to obtain the text features $h(s_i)$. Then the text features $h(s_i)$ could be used to construct classifier weights $W \in R^{d \times k}$, where $d$ is the dimension of text features. Finally, we combine the image encoder $g$ with the classifier weights $W$ to obtain a classification model $f(x)=g(x)^\mathrm{T}W$.
We empirically show that \textbf{CT} performs comparably with other tuning methods.
Compared with complex designed tuning methods, we hope to use a simpler method for better investigating the effectiveness of synthetic data.

\vspace{-0.1cm}
\subsection{Is Synthetic data ready for Zero-shot image recognition?} \label{sec: zero-shot}
\vspace{-0.1cm}

Our aim is to investigate to what degree synthetic data are beneficial to zero-shot tasks and how to better leverage synthetic data for zero-shot learning.

\noindent \textbf{Zero-shot Image Recognition.} 
We study the inductive zero-shot learning setting where no real training images of the target categories are available. 
CLIP models are pre-trained with large-scale image-caption pairs, and the similarities between paired image features (from an image-encoder $g$) and text features (from a text-encoder $h$) are maximized during pre-training. The pre-trained feature extractor can then be used to solve zero-shot tasks where given an image, its features from $g$ are compared with text features of different classes from $h$ and the image is further assigned to the class that has the largest similarity in the CLIP text-image feature space.

\noindent
\textbf{Synthetic Data for Zero-shot Image Recognition.}  
Though CLIP models exhibit strong zero-shot performance thanks to the large-scale vision-language dataset for pre-training, there are still several shortcomings when the model is deployed for a downstream zero-shot classification task, which may be attributed to unavoidable data noise in CLIP's pre-training data or the label space mismatch between pre-training and the zero-shot task. Hence, with a given label space for a zero-shot task, we study whether synthetic data can be used to better adapt CLIP models for zero-shot learning.

\textit{How to generate the data?} Given a pre-trained text-to-image generation model, to synthesize novel samples, the basic (\textbf{B}) strategy is to use the label names of the target categories to build the language input and generate a corresponding image. Then, the paired label names and synthesized data can be employed to train the classifier with the feature extractor frozen. 

\textit{How to enrich diversity?} Only using the label names as inputs might limit the diversity of synthesized images and cause bottlenecks for validating the effectiveness of synthetic data. Hence, we leverage an off-the-shelf word-to-sentence T5 model (pre-trained on ``Colossal Clean Crawled Corpus'' dataset \citep{raffel2020exploring} and finetuned on CommonGen dataset \citep{lin2019commongen}) to increase the diversity of language prompts and the generated images, namely language enhancement (\textbf{LE}), hoping to better unleash the potential of synthesized data.
Concretely, we input the label name of each class to the word-to-sentence model which generates diversified sentences containing the class names as language prompts for the text-to-image generation process. For example, if the class label is ``airplane", then the enhanced language prompt from the model could be ``a white airplane hovering over a beach and a city''. The enhanced text descriptions introduce rich context descriptions. 

\textit{How to reduce noise and enhance robustness?} 
It's unavoidable that the synthesized data may contain low-quality samples. This is even more severe in the setting with language enhancement as it may introduce undesired items into language prompts (see Figure A.2 in Appendix for visual examples). Hence, we introduce a CLIP Filter (\textbf{CF}) strategy to rule out these samples. Specifically, CLIP zero-shot classification confidence is used to assess the quality of synthesized data, and the low-confidence ones are removed.
Besides, as soft-target is more robust than hard-target in countering sample noise, we study whether soft cross-entropy loss (\textbf{SCE}, see Sec. C.4 in Appendix) which uses the normalized clip scores as a target could be used to enhance robustness against data noise.

\noindent
\textbf{Experiment Setup}.
We select 17 diverse datasets covering object-level (CIFAR-10 and CIFAR-100 (\citep{krizhevsky2009learning}, Caltech101 \citep{fei2006one}, Caltech256 \citep{griffin2007caltech}, ImageNet \citep{deng2009imagenet}), scene-level (SUN397 \citep{xiao2010sun}), fine-grained (Aircraft \citep{maji13fine-grained}, Birdsnap \citep{berg-birdsnap-cvpr2014}, Cars \citep{KrauseStarkDengFei-Fei_3DRR2013}, CUB \citep{wah2011caltech}, Flower \citep{nilsback2008automated}, Food \citep{bossard2014food}, Pets \citep{parkhi2012cats}), textures (DTD \citep{cimpoi2014describing}), satelite images (EuroSAT \citep{helber2019eurosat}) and robustness (ImageNet-Sketch \citep{wang2019learning}, ImageNet-R \citep{hendrycks2021many}) for zero-shot image classification. 
For synthetic data amount, we generate 2000 (study of synthetic image number in Appendix Sec. B.3) synthetic images for each class in \textbf{B} and \textbf{LE}. For \textbf{LE}, we generate 200 sentences for each class. 

\begin{table}[t]%[htbp] 
   \centering
    \vspace{-0.8cm}
   \begin{small}
       \setlength\tabcolsep{5pt} 
       \begin{tabular}{c|c|cc|cc}
           \bottomrule[1pt]
           Dataset & Task & CLIP-RN50 & CLIP-RN50+SYN & CLIP-ViT-B/16 & CLIP-ViT-B/16+SYN  \\ 
           \hline
           CIFAR-10 &o& 70.31 & 80.06 \gain{9.75}  & 90.80 & 92.37 \gain{1.57} \\
           CIFAR-100 &o & 35.35 & 45.69  \gain{10.34} & 68.22 & 70.71 \gain{2.49} \\
           Caltech101 &o & 86.09 & 87.74 \gain{1.65} & 92.98 & 94.16 \gain{1.18} \\
           Caltech256 & o &73.36 & 75.74 \gain{2.38} & 80.14 & 81.43 \gain{1.29}\\
           ImageNet &o & 60.33 & 60.78   \gain{0.45} & 68.75 & 69.16 \gain {0.41} \\
           SUN397 & s &58.51 & 60.07      \gain{1.56} & 62.51 & 63.79 \gain{1.28} \\
           Aircraft &f & 17.34 & 21.94   \gain{4.60} & 24.81 & 30.78 \gain{5.97} \\
           Birdsnap &f& 34.33 & 38.05   \gain{3.72} & 41.90 & 46.84 \gain{4.94}\\
           Cars &f & 55.63 & 56.93       \gain{1.30} & 65.23 & 66.86 \gain{1.63} \\
           CUB &f& 46.69 & 56.94        \gain{10.25} & 55.23 & 63.79 
          \gain{8.56}\\
           Flower &f & 66.08 & 67.05     \gain{0.97} & 71.30 & 72.60 \gain{1.30} \\
           Food &f & 80.34 & 80.35       \gain{0.01} & 88.75 & 88.83 \gain{0.08} \\
           Pets &f & 85.80 & 86.81       \gain{1.01} & 89.10 & 90.41 \gain{1.31} \\
           DTD & t & 42.23 & 43.19           \gain{0.96} & 44.39 & 44.92 \gain{0.53}\\
           EuroSAT &si & 37.51 & 55.37 \gain{17.86} & 47.77 & 59.86 \gain{12.09}\\

           ImageNet-Sketch & r & 33.29 & 36.55 \gain{3.26}& 46.20 & 48.47 \gain{2.27} \\
           ImageNet-R & r & 56.16 & 59.37 \gain{3.21} & 74.01 & 76.41 \gain{2.40}\\
           
           \hline
           Average & / & 55.13 & 59.47 \gain{4.31} & 65.42 & 68.32 \gain{2.90}\\
           \toprule[0.8pt] 
       \end{tabular}
   \end{small}
   \vspace{-0.4cm}  
   \caption{\textbf{Main Results on Zero-shot Image Recognition.} All results are top-1 accuracy on test set. o: object-level. s: scene-level. f: fine-grained. t: textures. si: satellite images. r: robustness. }
    \vspace{-0.6cm}
   \label{tab: zsl main}
\end{table}

\noindent
\textbf{Main Results:} 
 1) zero-shot classification results on 17 datasets; 2) study of synthetic data diversity; 3) study of synthetic data reliability; 4) study of model/classifier tuning; 5) study of the behavior of synthetic data for zero-shot classification in the training from scratch settings.

\textit{Synthetic data can significantly improve the performance of zero-shot learning.} Our main studies in zero-shot settings are conducted with CLIP-RN50 (ResNet-50 \citep{he2016deep} and CLIP-ViT-B/16 (ViT-B/16 \citep{dosovitskiy2020image}) as CLIP backbone), and we report results with our best strategy of \textbf{LE+CF+SCE}. As shown in Table \ref{tab: zsl main}, on 17 diverse downstream zero-shot image classification datasets, we achieve a remarkable average gain of 4.31\% for CLIP-RN50 and 2.90\% for CLIP-ViT-B/16 in terms of top-1 accuracy. Significantly, on the EuroSAT dataset, we achieve the largest performance boost of 17.86\% for CLIP-RN50 in top-1 accuracy. We notice that the performance gain brought by synthetic data varies differently across datasets, which is mainly related to GLIDE’s training data distribution. The training data distribution of the text-to-image generation model GLIDE would exhibit bias and produce different domain gaps with different datasets (see Sec. A.2 in Appendix for more analysis).

\begin{table*}[t]%[htbp]
   \centering
   \begin{small}
   \vspace{-0.95cm}
      \setlength\tabcolsep{1.6pt} 
       \begin{tabular}{c|c|c|c|c|c|c|c}
           \bottomrule[1pt]
           \multirow{2}{*}{Dataset} & \multirow{2}{*}{CLIP}  & \multicolumn{2}{c|}{B} & \multicolumn{2}{c|}{LE} & \multicolumn{2}{c}{LE+CF} \\
           \cline{3-8} 
           &  & CE & SCE & CE & SCE & CE & SCE\\
           \hline
           CIFAR-10 & 70.31  & 77.39 \gain{7.08} & 78.23 \gain{7.92} & 77.20 \gain{6.89} & 77.55 \gain{7.24} & 80.01 \gain{9.70} & \highlight{80.06 \gain{9.75}} \\
           CIFAR-100 & 35.35 & 43.99 \gain{8.64} & 44.25 \gain{8.90} & 44.08 \gain{8.73} & 44.91 \gain{9.56} & 44.55 \gain{9.20} & \highlight{45.69 \gain{10.34}} \\
           EuroSAT & 37.51 & 45.64 \gain{8.13} & 48.23 \gain{10.72} &
           53.26 \gain{15.75} & 54.94 \gain{17.43} & 54.75 \gain{17.24} & \highlight{55.37 \gain{17.86}}\\
           \toprule[1pt]
         %   \bottomrule[1pt]
         \end{tabular}
      \end{small}
      \vspace{-0.2cm}
      \caption{Ablation study on \textbf{Language Enhancement (LE), CLIP-based Filtering (CF), and Soft-target Cross-Entropy (SCE).}}
      \vspace{-0.4cm}
      \label{tab: zsl ablation}
  \end{table*}

\textit{Language diversity matters.} 
By introducing more linguistic context into the text input, \textbf{LE} helps increase the diversity of synthetic data.
As shown in Table \ref{tab: zsl ablation}, \textbf{LE} can achieve additional performance gains upon \textbf{B} in most cases (0.66$\uparrow$ on CIFAR-100, 6.71$\uparrow$ on EuroSAT), which demonstrates the efficacy of \textbf{LE} and the importance of synthetic data diversity for zero-shot classification. 

\textit{Reliability matters.} 
 While \textbf{LE} could help increase the diversity of synthetic data, it also introduces the risks of noisy samples.
 Observed on CIFAR-10 in Table \ref{tab: zsl ablation}, \textbf{LE} sometimes even brings performance drops compared with \textbf{B} (0.68$\%\downarrow$ on CIFAR-10), which may attribute to the noise introduced by enhanced language prompts, {\eg} the sentence extended from the class name word may contain other class names or confusing objects. 
 Fortunately, with \textbf{CF} to filter out unreliable samples, \textbf{LE+CF} yields consistent improvement upon \textbf{B}. Moreover, \textbf{SCE} generally achieves better performance than \textbf{CE}, showing its better adaptation to label noise.
 
%   \textit{add a subtitle for this part??} 
  \textit{Classifier tuning is enough for CLIP, while tuning with the pre-trained encoder leads to degradation, mainly due to domain gaps.}
  Here, we investigate if only tuning the final classifier is the optimal solution in our setting with synthetic data. As shown in Table \ref{tab: zsl param tuned}, we tune different proportions of the full model parameters 
  %(we do not count in the parameters of the text encoder) 
  on synthetic data for EuroSAT (0.02\% corresponds to our default case where only the classifier is tuned), and report the zero-shot performance on the test set of EuroSAT. The best results are obtained by only tuning the classifier, and the performance gradually decreases as we gradually incorporate more parameters in the encoder for optimization, which agrees with the traditional strategy. 
  For understanding why synthetic data may harm pre-trained image encoder, we experiment with real-world data with domain shifts and find they behave similarly to synthetic data (Appendix Sec. B.2), which suggests that domain gap is the main reason for the phenomenon. We argue that synthetic data might have a better chance to overcome domain shifts in comparison with real-world data since we can customize and keep the label space of the synthetic data in line with the down-stream dataset and use strategies during synthesizing to alleviate domain shifts.

  \begin{table}[htbp] 
   \centering
   \vspace{-0.2cm}
   \begin{small}
       \setlength\tabcolsep{9pt} 
       \begin{tabular}{ccccccccc}
           \bottomrule[1pt]
           Param Tuned (\%)& 0 & 0.02 & 0.04 & 62.50  & 64.06 & 69.53 & 82.81 & 92.19   \\ 
           \hline
           Acc & 37.51 & \highlight{55.37} & 55.11 & 55.28 & 54.56 & 54.34 & 53.63 & 52.09  \\
           \toprule[0.8pt] 
       \end{tabular}
   \end{small}
   \vspace{-0.3cm}  
   \caption{\textbf{Parameters tuned v.s. Accuracy.} Dataset: EuroSAT.}
   \vspace{-0.3cm}
   \label{tab: zsl param tuned}
\end{table}

  \begin{table}[htbp] 
   \centering
    \vspace{-0.2cm}
   \begin{small}
       \setlength\tabcolsep{10pt} 
       \begin{tabular}{ccccccccc}
           \bottomrule[1pt]
           Real shot & 1 & 16 & 32  & 64 & 80 & 90 & 95 & 100  \\ 
           \hline
           Acc &  2.48 & 10.4 & 14.95 & 21.96 & 24.4 & 25.52 & 27.99 & 29.95 \\
           \toprule[0.8pt] 
       \end{tabular}
   \end{small}
   \vspace{-0.2cm}  
   \caption{\textbf{Setting when training from scratch.} Dataset: CIFAR-100. }
    \vspace{-0.2cm}
   \label{tab: zsl from scrach}
\end{table}

\textit{Synthetic data deliver inferior performance in the training from scratch setting and are much less data-efficient than real data.} 
To exclude the influence of powerful CLIP initialization in our study of synthetic data, we also conduct a from-scratch setting on the CIFAR-100 dataset, where we optimize a ResNet-50 model from random initialization.
Given the label space of the CIFAR-100 dataset, we generate a synthetic dataset of 50k (500 images per class) to train a ResNet-50 model from scratch for image classification. We achieve a performance of 28.74\% top-1 accuracy {on CIFAR-100 test set}, which is much lower than the performance of the pre-trained CLIP model (see Table \ref{tab: zsl main}).
%Further, we hope to figure out our synthetic data equal to how many real in-domain training data. 
This might be attributed to the quality and diversity of data. The CLIP model benefits from diverse real-world data.
% \xjqi{an analysis is needed here to explain why the above happens. CLIP large-scale pre-trained, quality??}
Further, we hope to investigate how many real in-domain training data can match the performance of our 50k synthetic data. As shown in Table \ref{tab: zsl from scrach},
training with 95 images per category ($95 \times 100 = 9.5$k) will achieve comparable performance as that of 50k synthetic data.
%Training on our synthetic data (around 50k) from scratch leads to a zero-shot performance of 28.74\% top-1 accuracy, which matches the performance (27.99\%) of training on around 95 shot (95*100=9.5k) real in-domain CIFAR-100 training samples as shown in Table \ref{tab: zsl from scrach}. 
This manifests that synthetic data are not as efficient and effective as real data when solving downstream tasks. It requires around 5 times more data in order to achieve a comparable performance as that of real data.
Note that we find further increasing the amount of synthetic data will not deliver further performance gains for the downstream classification task. We expect that further investigations on synthesis quality will bring new opportunities in this area which will be our future work. 

\noindent \textbf{Summary.}
Current synthetic data from text-to-image generation models could indeed bring significant performance boosts for a wide range of zero-shot image classification tasks, and is readily applicable with carefully designed strategies such as large-scale pre-trained models. 
Diversity and reliability matter for synthetic data when employed for zero-shot tasks. 
When the model is trained from scratch with synthetic data, synthetic data cannot deliver satisfactory performance and are much less data-efficient and effective for solving the classification task in comparison with real data.

\vspace{-0.2cm}
\subsection{Is Synthetic data ready for Few-shot Image Recognition?} \label{sec: few-shot}
\vspace{-0.1cm}

\noindent
In this section, we explore the effectiveness of synthetic data for few-shot tasks and how synthetic data impact the performance as more and more shots are included. Also, we design effective strategies to better leverage synthetic data.

\textbf{Few-shot Image Recognition.}
We adopt the CLIP-based method as the model for few-shot image recognition due to its state-of-the-art performance \citep{radford2021learning}. As discussed previously, various prompt learning based methods can be treated as tuning the classifier weights. We thus study how to tune the classifier weights with synthetic data.  
In an N-way M-shot case, we are given M real images of each test class, where M $\in \{\text{1, 2, 4, 8, 16}\}$ in our experiments. With a total of $\text{N}\times\text{M}$ training samples, we hope to achieve favorable performance on a hold-out test set of the N classes.

\textbf{Synthetic Data for Few-shot Image Recognition.}  
While there have been a few attempts to study how to better adapt CLIP models for few-shot tasks \citep{zhou2022learning, zhou2022conditional, zhang2022tip}, they all focus on the model optimization level, and none have explored from the data level. Here, we systematically study whether and how synthetic data can be employed for solving few-shot image recognition tasks. 

With the experience from synthetic data for zero-shot tasks, we adopt the best strategy ({\ie} \textbf{LE+CF}) in the zero-shot setting as the basic strategy (\textbf{B}). Further, as the few-shot real samples can provide useful information on the data distribution of the classification task, we develop two new strategies leveraging the in-domain few-shot real data for better using synthetic data: 1) Real Filtering (\textbf{RF}): given synthetic data of one class $c$, we use the features of few-shot real samples to filter out synthetic images whose features are very close to the features of real samples that belong to other categories different from class $c$; %, namely  Real Filtering ; 
%that using M shot images of each class to perform feature similarity filtering, filtering out the images in each class that have larger feature similarities to other classes; 
2) Real guidance (\textbf{RG}): we use the few-shot real samples as guidance to generate synthetic images where the few-shot real samples (added noise) replace the random noise at the beginning of the generation to guide the diffusion process (details in Appendix Sec. C.3).

\noindent
\textbf{Experiment Setup}.
For datasets, we carefully select 8 image classification datasets from recent works \citep{zhou2022learning, zhou2022conditional, zhang2022tip} that conduct few-shot learning upon CLIP: ImageNet \citep{deng2009imagenet}, Caltech101 \citep{fei2006one}, Pets \citep{parkhi2012cats}, Cars \citep{KrauseStarkDengFei-Fei_3DRR2013}, Aircraft \citep{maji13fine-grained}, SUN397 \citep{xiao2010sun}, DTD \citep{cimpoi2014describing}, EuroSAT \citep{helber2019eurosat}. 
For synthetic image number, we generate 800  (study of synthetic image number in Appendix Sec. B.3) images per class for \textbf{RG} method to approximately match the number of images in \textbf{B} and \textbf{RF}.

\noindent
\textbf{Main Results:}
 1) few-shot classification results on 8 datasets; 2) ablation study of training strategy; 3) ablation study of synthetic data generation strategy; 4) ablation study of BN strategy.

\textit{Synthetic data can boost few-shot learning and the positive impact of synthetic data will gradually diminish with the increase of real data shots.}
As shown in Figure \ref{fig: few-shot} (results of more datasets are in the Appendix Sec. B.1), with only few-shot real images for training, our implemented \textbf{CT w. init} (classifier weights initialized from CLIP text embeddings)
performs comparably with the state-of-the-art CLIP tuning methods \textbf{Tip Adapter} \citep{zhang2022tip} and \textbf{CoOp} \citep{zhou2022learning}. 
\textbf{CT w. Syn} represents our results of applying synthetic data with mix training, real image as guidance, and freezing BN strategies.
With the help of generated synthetic data, \textbf{CT w. Syn} achieves noticeable performance gains upon \textbf{CT w. init}, and achieves a new state-of-the-art few-shot learning performance across different datasets. We argue that for data-scarce few-shot classification, synthetic data could help address the insufficient data problem to boost performance. However, we notice that the boost from synthetic data gradually diminishes as the real shot number increases. 
We state that the effectiveness of each sample in real data is high since there's no domain gap; in contrast, synthetic data suffer from domain gaps and perform less efficiently. In addition, the positive effects of the few-shot real data may overlap with that of synthetic data. Thus, with the increase of real data, the overlapping becomes serious and the positive impacts of synthetic data are reduced.

\begin{figure} 
\vspace{-0.95cm}
   \begin{center}
   % \fbox{\rule{0pt}{2in} \rule{.9\linewidth}{0pt}} 
   \includegraphics[width=1.0\linewidth]{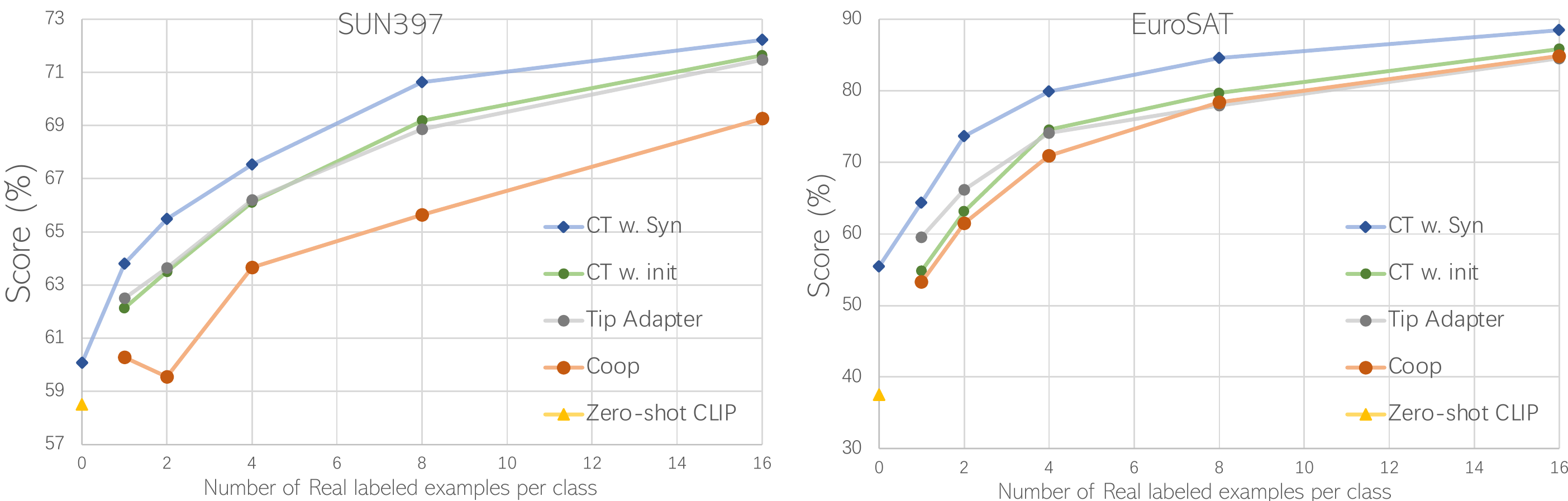} 
   \end{center}
    \vspace{-0.4cm}
      \caption{Results for few-shot image recognition. Results on all 8 datasets are provided in Appendix. }
   \label{fig: few-shot} 
   \vspace{-0.5cm} 
   \end{figure}

\textit{Mix Training fits few-shot learning with synthetic data.}
Now that we have two parts of data, {\ie} few-shot real data and synthetic data, we could either 1) \textbf{phase-wise} train on each part of data with two training phases, or 2) adopt \textbf{mix training} that simultaneously utilizes two parts of data to update the model in each iteration. Details of phase-wise/mix training in Appendix Sec. C.5.2. We provide the results in Table \ref{tab: few-shot training strategy}: we study on the EuroSAT dataset and use synthetic data generated from the \textbf{RG} method; under different shot number settings, mix training performs consistently better than two phase-wise strategies. We suggest that mix training could help learn better classifiers since each part could function as a regularization for the other: synthetic data help alleviate instabilities brought by limited real samples, and real data help address the noise and domain gap of synthetic data.

\textit{Employing real data as guidance can alleviate domain differences and boost performance.} 
We compare three strategies of synthetic data generation for few-shot tasks. As shown in Table \ref{tab: few-shot syn data strategy}, both \textbf{RF} and \textbf{RG} provide performance gains upon \textbf{B} which is the best strategy in the zero-shot setting. This demonstrates the importance of utilizing the domain knowledge from few-shot images for preparing the synthetic data. Further, \textbf{RG} significantly outperforms \textbf{RF}, yielding the best performance. 
This shows utilizing real data as guidance of the diffusion process help reduce the domain gap (visual illustrations in the Appendix Sec. B.8). 

\begin{table}[htbp] 
   \centering
%    \vspace{-0.2cm}
   \begin{minipage}{0.24\textwidth}
       \tablestyle{3.6pt}{1.2}
       \begin{tabular}{ccc}
           \bottomrule[1pt]
           B & RF & RG \\ 
           \hline
           87.1 & 87.33 & 88.47 \\
           \toprule[0.8pt] 
       \end{tabular}
   \vspace{-0.01cm}  
   \caption{Ablation for Basic strategy (\textbf{B}), Real Filtering (\textbf{RF}), Real Guidance (\textbf{RG}) on EuroSAT, 16 shot. }
   \label{tab: few-shot syn data strategy}
   \vspace{-0.25cm}
   \end{minipage}
   \hfill
   \begin{minipage}{0.36\textwidth}
      \tablestyle{2.6pt}{1.2}
      \begin{tabular}{ccc}
           \bottomrule[1pt]
           Train data & Freeze BN? & Test Acc \\ 
        %   \hline
        %   / & / & 37.51 \\ 
           \hline
           Real &  & 75.31 \\
           Real & \checkmark & \highlight{85.63} \\
           \hline
           Syn & & 44.73 \\
           Syn & \checkmark & \highlight{55.37} \\
           \toprule[0.8pt] 
       \end{tabular}
   \vspace{-0.01cm}  
   \caption{\textbf{Frozen BN works better} for 16-shot settings on EuroSAT.}
   \label{tab: few-shot BN strategy}
   \vspace{-0.25cm}
   \end{minipage}
   \hfill
%    \vspace{-0.2cm}
   \begin{minipage}{0.36\textwidth}
   \scalebox{0.9}{
   \centering
   \tablestyle{1.6pt}{1.0}
       \begin{tabular}{cccc}
           \bottomrule[1pt]
           \multirow{2}{*}{M-shot}  & \multicolumn{2}{c}{Phase-wise} & \multirow{2}{*}{\makecell[c]{Mix\\training}} \\ 
             & syn $\rightarrow$ real & real $\rightarrow$ syn &  \\ 
           \hline
            1&63.01&63.32&\highlight{64.36} \\
            2&72.24&72.85&\highlight{73.62} \\
            4&78.88&79.21&\highlight{79.88} \\
            8&83.64&83.99&\highlight{84.57} \\
           16&87.10&87.44&\highlight{88.47} \\
           \toprule[0.8pt] 
       \end{tabular}
       }
   \vspace{-0.02cm}  
   \caption{\textbf{Mix training works better} for few-shot tasks on EuroSAT.}
   \vspace{-0.25cm}
   \label{tab: few-shot training strategy}
   \end{minipage}
\end{table}

\textit{Frozen BN works better.}
Lastly, we investigate batch normalization (BN) strategies for our few-shot settings with synthetic data. As shown in Table \ref{tab: few-shot BN strategy}, for both real and synthetic data, freezing the BN layers yields much better performance. We analyze that for real data, it is hard to get a good estimation of BN statistics when the number of images is limited. As for synthetic data, we attribute this to the statistical difference between different domains. Hence, we freeze BN layers during tuning for few-shot settings.

\noindent \textbf{Summary.} Synthetic data from text-to-image generation models could readily benefit few-shot learning and achieve a new state-of-the-art few-shot classification performance with strategies we present in this paper. However, the positive impact of synthetic data will diminish as more shots of real data are available which further confirms our previous claim that synthetic data are still not as effective as real data in training classification models.

\vspace{-0.25cm}
\subsection{Is Synthetic data ready for Pre-training?} \label{sec: pre-train}
\vspace{-0.2cm}

Finally, we study whether synthetic data are effective in large-scale pre-training.
We also present effective strategies to better leverage synthetic data for model pre-training.

\textbf{Pre-training for Transfer Learning.} 
Recently, it has become a common practice to first pre-train models on large-scale datasets to obtain a well-trained feature extractor and then fine-tune the models on downstream tasks with labeled data ({\ie} transfer learning). There have been various successful pre-training methods, including supervised pre-training
\citep{joulin2016learning, li2017learning, mahajan2018exploring, sun2017revisiting, kolesnikov2020big}, self-supervised pre-training \citep{chen2020simple, he2020momentum, caron2020unsupervised, grill2020bootstrap, chen2021exploring, zbontar2021barlow, ye2019unsupervised}, and semi-supervised pre-training \citep{xie2020self, pham2021meta}. 

\textbf{Synthetic data for Pre-training.}
Since data amount and diversity play important roles in pre-training, 
% rather than data cleaning \xjqi{why not data cleaning}, 
we adopt the synthetic data generation strategy \textbf{LE} solely to maximize the scale of synthetic pre-training data.
We study two settings for generating synthetic data for pre-training: 1) downstream-aware, where we have access to the label space of the downstream task, and thus we generate synthetic data according to the label space of the downstream task; 2) downstream-agnostic, where we have no access to downstream tasks in the pre-training stage, and we turn to a relatively general and diverse label space such as ImageNet-1K. For pre-training methods, we experiment with supervised pre-training and self-supervised pre-training methods.

% \noindent
% \textbf{Method}: 
% Since pre-training values much more of data amount and diversity rather than data cleaning, we adopt the synthetic data generation strategy \textbf{LE} solely. We generate a labeled pre-training dataset from a certain label space (\eg ImageNet-1K label space or downstream label space), and either perform supervised pre-training or discard the labels and use self-supervised pre-training. We compare the synthetic pre-trained models with random initialization and ImageNet-1K pretrained models by transfering to downstream tasks.

\noindent
\textbf{Experiment Setup}.
We compare synthetic pre-trained models with models of random initialization and models of ImageNet-1K pre-training in terms of their transfer learning abilities. 
For downstream-aware settings: we conduct supervised pre-training on synthetic data generated according to CIFAR-100 label space and then transfer to CIFAR-100 through finetuning for evaluation. 

For downstream-agnostic settings: we perform supervised pre-training and self-supervised pre-training (we adopt Moco v2 \citep{chen2020improved} framework for its simplicity and reproducibility) on synthetic data generated from ImageNet-1K label space and evaluate the transfer performance by finetuning the pretrained models on a object detection dataset --  PASCAL VOC \citep{everingham2010pascal}. 
Further, we experiment with ImageNet-2K label space (original ImageNet-1K and another non-overlapping 1K label names randomly selected from ImageNet-21K) to study the factors of data diversity and amount in synthetic pre-training. 
We use ResNet-50 as the default backbone when not else noted, and also experiment with a ViT-based backbone, {\ie} DeiT-S \citep{touvron2021training}.

\noindent
\textbf{Results for Downstream-aware settings}. 
% For downstream-aware synthetic pre-training targeted on CIFAR-100, 
We generate synthetic data of different sizes from CIFAR-100 label space, {\ie} {1$\times$, 2$\times$, 3$\times$} ImageNet-1K data size, concretely {1.2M, 2.4M, 3.6M}. We pre-train the model on the generated synthetic labeled set in a supervised manner, and then perform evaluation after finetuning the model on CIFAR-100. As shown in Table \ref{tab: pre-train CIFAR-100}, %with only 1$\times$ amount of ImageNet-1K dataset size synthetic data (83.90), 
with an equivalent amount of data as that of ImageNet-1K (1.2M), synthetic
 data for pre-training can largely reduce the gap between training from scratch (78.83\%) and ImageNet- pre-trained model (84.50\%).
Moreover, with 2$\times$ and 3$\times$ synthetic data, pre-training on synthetic data outperforms ImageNet-1K pre-training with a noticeable margin. 
%Further, with 2$\times$ and 3$\times$ synthetic data, we outperform ImageNet-1K pre-training with noticeable margins.
In addition, when we initialize the model from ImageNet-1K pre-trained weights and  pre-train the model on synthetic data, we obtain extra boosts upon both results. 

We conclude that for downstream-aware synthetic pre-training, synthetic data deliver close performance as that of ImageNet-1K pretraining with the same amount of data, synthetic data amount helps improve the results to outperforming ImageNet-1K pre-training, and synthetic pre-training could further benefit from ImageNet-1K pre-training.

\textbf{Results for Downstream-agnostic settings}.
% For downstream-agnostic synthetic pre-training, w
We first experiment with ImageNet-1K label space with 1$\times$ or 2$\times$ ImageNet-1K data size, {\ie} 1.2M/2.4M IN-1K Syn. We perform supervised pre-training and self-supervised pre-training ({\ie} Moco v2) on the generated synthetic data, and evaluate the pre-training results by transferring to the CIFAR-100 image classification task or the PASCAL VOC detection task. As it is too costly to validate all settings ({\eg}, it takes more than 1 week to train Moco v2 on 4.0M synthetic data), we select several representative settings of interest to validate the effectiveness of synthetic data without hurting our conclusion.

As shown in Table \ref{tab: pre-train imgnet super} and \ref{tab: pre-train imgnet moco}, with 1.2M IN-1K Syn, both supervised pre-training (79.00\%) and self-supervised pre-training (81.55\%) could largely approach their IN-1K Real counterparts (super.:81.3\%; self-super.:82.44\%) and largely outperforms the result without pre-training (66.08\%). When increasing the data amount to 2.4M, the transferred results further increase, and the unsupervised pre-training method, {\ie} Moco v2, performs better in utilizing our synthetic data thanks to its independence of labels, yielding a 82.13\% transferred performance which surpasses supervised pre-training on IN-1K Real (81.30\%) and is on par with its Moco v2 counterpart at IN-1K Real (82.44\%). 
Next, we expand the label space by adding another 1K categories, producing IN-2K Syn. The enlarged diversity and data amount further bridge the gap between synthetic pre-training results and IN-1K Real pre-training results. Noticeably, the unsupervised pre-trained model Moco v2 (82.29\%) largely approaches the IN-1K Real counterpart (82.44\%) with negligible performance drop of 0.15\%.
Furthermore, when initialized from IN-1K Real pre-trained weights, both supervised and self-supervised pre-training improve upon both pure real data and synthetic data for pre-training.

While the above results are all obtained with convolutional-based backbone {\ie} ResNet50, we further explore with a recent ViT-based backbone {\ie} DeiT-S \citep{touvron2021training}. Surprisingly, ViT-based backbone is shown to be more advantageous compared with convolution-based backbone for synthetic pre-training: outperforming ImageNet pre-training results in the downstream-agnostic settings. Equipped with ViT-based backbone, on only 1.2M IN-1K synthetic data, we achieve comparable performance  (87.98\%) with ImageNet pre-training (88.07\%). Further increasing the data amount (88.39\%) and label space (88.57\%, 88.91\%) of pre-training data leads to higher performance than ImageNet pre-training. 
ViT-based backbones have stronger ability for learning from large-scale data and are more robust \citep{pinto2021vision}, and thus could better benefit from synthetic pre-training where data are more noisy and data scale could be easily increased.
\begin{table}
   \begin{minipage}{0.48\textwidth}
    \vspace{-0.75cm}
   \scalebox{0.93}{
   \centering
      \tablestyle{2.6pt}{1.38}
       \begin{tabular}{c|c|cccc}
           \bottomrule[1pt] 
           \multirow{2}{*}{\makecell[c]{Data}} & \multirow{2}{*}{\makecell[c]{pre-trained \\on IN-1k?}}& \multicolumn{4}{c}{Syn. images amount}\\
           & & 0 & 1.2M & 2.4M & 3.6M \\
           \hline
           (None) & & 78.83 & - & - & - \\
        %   IN-1K Real & 81.30 & - & - & 82.44 & - & -  \\
           C100 Syn &  & - & 83.90 & \highlight{85.03} & \highlight{85.24} \\
           \hline
           (None) & \checkmark & 84.50 & - & - & - \\
           \hline
           C100 Syn &  \checkmark & -& \highlight{84.90} & \highlight{85.32} & \highlight{85.52} \\
           \toprule[1pt]
         %   \bottomrule[1pt]
         \end{tabular} 
         }
         \vspace{-0.2cm}  
         \caption{ Results on CIFAR-100 with \textbf{downstream-aware supervised pre-training}. C100: CIFAR100.}
         \vspace{+0.1cm}  
         \label{tab: pre-train CIFAR-100}
      \end{minipage}
      \hfill
        \begin{minipage}{0.48\textwidth}
         \vspace{-0.75cm}
      \scalebox{0.93}{
      \centering
      \tablestyle{2.6pt}{1.38}
      \begin{tabular}{c|c|cccc}
           \bottomrule[1pt] 
           \multirow{2}{*}{\makecell[c]{Data}} &\multirow{2}{*}{\makecell[c]{pre-trained \\on IN-1k?}}& \multicolumn{4}{c}{Syn. images amount}\\
           & & 0 & 1.2M & 2.4M & 4.0M  \\
      \hline
      (None) & & 69.29 & - & - & - \\
      IN-1K Syn & &  - & 87.98 & \highlight{88.39} & - \\
      IN-2K Syn & &  - & - & \highlight{88.57} & \highlight{88.91} \\
      \hline
      (None) & \checkmark & - & 88.07 & - & -  \\
    %   IN-1K Real + IN-2K Syn & - & - & -  \\
           \toprule[1pt]
         %   \bottomrule[1pt]
         \end{tabular} 
         }
         \vspace{-0.2cm}  
         \caption{Results on CIFAR-100 with \textbf{downstream-agnostic supervised pre-training}. Backbone: DeiT-S. } 
         \label{tab: pre-train imgnet deit}
        %  \vspace{-0.1cm}  
      \end{minipage}
   \hfill
   \begin{minipage}{0.48\textwidth}
   \scalebox{0.93}{
   \centering
      \tablestyle{2.6pt}{1.2}
      \vspace{+0.3cm} 
       \begin{tabular}{c|c|cccc}
           \bottomrule[1pt] 
           \multirow{2}{*}{\makecell[c]{Data}} & \multirow{2}{*}{\makecell[c]{pre-trained \\on IN-1k?}}& \multicolumn{4}{c}{Syn. images amount}\\
           & & 0 & 1.2M & 2.4M & 4.0M  \\
           \hline
           (None) & & 66.08 & - & - & - \\
        %   IN-1K Real & 81.30 & - & - & 82.44 & - & -  \\
           IN-1K Syn &  & - & 79.00 & 80.00 & - \\
           IN-2K Syn &  & - & - & 80.54 & 80.72 \\
           \hline
           (None) & \checkmark & 81.30 & - & - & - \\
           \hline
           IN-1K Syn & \checkmark & -& - & \highlight{81.78} & - \\
           IN-2K Syn & \checkmark & -& - & \highlight{81.87} & \highlight{81.91} \\
           \toprule[1pt]
         %   \bottomrule[1pt]
         \end{tabular} 
         }
         \vspace{-0.2cm}  
         \caption{Results for object detection on PASCAL VOC with \textbf{downstream-agnostic supervised pre-training}, all results are reported in AP$_{50}$.}
         \vspace{-0.5cm}  
         \label{tab: pre-train imgnet super}
      \end{minipage}
      \hfill
      \begin{minipage}{0.48\textwidth}
      \scalebox{0.93}{
      \centering
      \tablestyle{2.6pt}{1.38}
      \begin{tabular}{c|c|cccc}
           \bottomrule[1pt] 
           \multirow{2}{*}{\makecell[c]{Data}} & \multirow{2}{*}{\makecell[c]{pre-trained \\on IN-1k?}}& \multicolumn{4}{c}{Syn. images amount}\\
           & & 0 & 1.2M & 2.4M & 4.0M  \\
      \hline
      (None) &  & 66.08 & - & - & - \\
      IN-1K Syn &  & - & 81.55 & 82.13 & - \\
      IN-2K Syn &  & - & - & 82.22 & 82.29 \\
      \hline
      (None) &\checkmark & 82.44 & - & - & -  \\
      \hline
      IN-1K Syn & \checkmark & - & - & \highlight{82.47} &-      \\
    %   IN-1K Real + IN-2K Syn & - & - & -  \\
           \toprule[1pt]
         %   \bottomrule[1pt]
         \end{tabular} 
         }
         \vspace{-0.2cm}  
         \caption{Results for object detection on PASCAL VOC with \textbf{downstream-agnostic self-supervised pre-training (Moco v2)}, all results are reported in AP$_{50}$.  }
         \label{tab: pre-train imgnet moco}
         \vspace{-0.5cm}  
      \end{minipage}
      \label{tab: pre-train agnostic, voc}
  \end{table}

\noindent
\textbf{Conclusion}. In terms of transfer abilities,  synthetic data from text-to-image generation models show surprisingly promising results for model pre-training, which is comparable to the standard ImageNet pre-training. We conclude our findings as follows:

% \begin{enumerate}[itemsep=0pt,topsep=0pt,parsep=0pt, leftmargin=13pt, partopsep=0pt]
\vspace{-0.2cm}
1. Data amount has positive impacts on synthetic pre-training; performance could be improved by increasing synthetic data size, but would gradually saturate as the amount of data increases.
\vspace{-0.2cm}

2. Synthetic data for pre-training is orthogonal to real data for pre-training.
\vspace{-0.2cm}

3. For downstream-aware synthetic pre-training, we significantly outperform IN-1K Real (1.2M) pre-training with 2.4M/3.6M synthetic data on CIFAR-100.
\vspace{-0.2cm}

4. For downstream-agnostic synthetic pre-training, we achieve comparable results with ImageNet (IN-1k) Real pre-training; self-supervised pre-training performs better than supervised pre-training, and ViT-based backbone performs better than convolutional-based backbone. Besides, increasing the label space size could further improve the performance.
% \end{enumerate}

% \noindent \textbf{Conclusion.}
\vspace{-0.3cm}
\section{Conclusion}
\vspace{-0.25cm}
We systematically investigate whether synthetic data from current state-of-the-art text-to-image generation models are readily applicable for image recognition. Our extensive experiments demonstrate that synthetic data are beneficial for classifier learning in zero-shot and few-shot recognition, bringing significant performance boosts and yielding new state-of-the-art performance. Further, current synthetic data show strong potential for model pre-training, even surpassing the standard ImageNet pre-training. We also point out limitations and bottlenecks for applying synthetic data for image recognition, hoping to arouse more future research in this direction.

%\section{Limitations} \label{sec: dis few}
\vspace{-0.15cm}
\textbf{Limitations.} In all investigated settings, we observe improved performance as the data amount and diversity (label space) increases. However, due to our limited computational resource, we are not able to further scale up data amount, which may take months to train one model. Besides, we are also not able to investigate larger model sizes and advanced architectures in the current investigation which is also worth exploring in the future. We present more discussions on limitations and future directions in the appendix.

\bibliographystyle{iclr2023_conference}

\clearpage

\renewcommand{\thetable}{A.\arabic{table}}
\renewcommand{\thefigure}{A.\arabic{figure}}

\appendix

\section{More Analysis}
\subsection{Limitations and future directions} \label{sec: limitation A}
\textit{A systematic way to study the language enhancement process is needed.}
In our experiments, we found language enhancement to be an effective way for enlarging synthetic data diversity. In this paper, we adopt an off-the-shelf word-to-sentence language model for this job, but this may bring certain risks and generate a high volume of noisy data. Therefore, effective constraints should be considered for language enhancement, which we leave as future work.

\textit{Larger synthetic data sizes for zero/few-shot tasks should be studied.}
We only generate synthetic data of median size for zero/few-shot tasks due to our limited computational resources; we encourage future works to further explore whether increasing data diversity and data amount could benefit these data-scarce tasks.

\textit{Co-training down-stream predictors with generative models.} In this paper, we propose a pre-trained off-line image generator to construct our synthetic training set. It would be interesting to investigate how to generate in-domain images by tuning the image generator together with the lateral model for down-stream tasks. This could potentially be useful for domain generalization.

\textit{Increasing scales for synthetic data and model capability for pre-training settings.} We observe improved performance when the data amount and diversity (label space) increase. However, due to our limited computational resource, we cannot further scale up the investigation, which may take months to train one model.  We hope our promising results could inspire future research to explore enlarged scales for synthetic data and model size (larger model capability should be better at benefiting from a larger amount of data) for pre-training settings.

\subsection{Analysis of differences of performance boost on different datasets in zero-shot settings
}
 \textit{The main reason is related to GLIDE’s training data distribution.} The training data distribution of the text-to-image generation model GLIDE would exhibit bias and produce different domain gaps with different datasets. While we do not have direct access to GLIDE’s training data, we may use synthesized images from GLIDE to explore. However, how to measure data and its relation to the performance for different tasks in a quantitative manner is still an open problem to our best knowledge as different task may have different levels of difficulties and data will impact the performance in different manners. An empirical thought is related to the synthesized image quality (e.g., synthesized images for DTD dataset is relatively noisy as shown in Figure \ref{fig: syn datasets}, where we only achieve 0.96\% gains), domain gaps (synthesized images are usually object-centric and exhibit large domain difference with scene-level dataset SUN397, where only 1.56\% boost achieved), and task difficulties (e.g., ImageNet is of high difficulty where only 0.45\% gains are obtained).

For further improving the results, we suggest that it would be interesting to investigate how to generate in-domain images by tuning the image generator together with the lateral model for downstream tasks. This could potentially better reducing domain gap to the downstream task.

\section{More Experiments}

\subsection{Main results on Few-shot tasks} \label{sec: few-shot A}
\begin{figure} 
    \begin{center}
    % \fbox{\rule{0pt}{2in} \rule{.9\linewidth}{0pt}} 
    \includegraphics[width=1.0\linewidth]{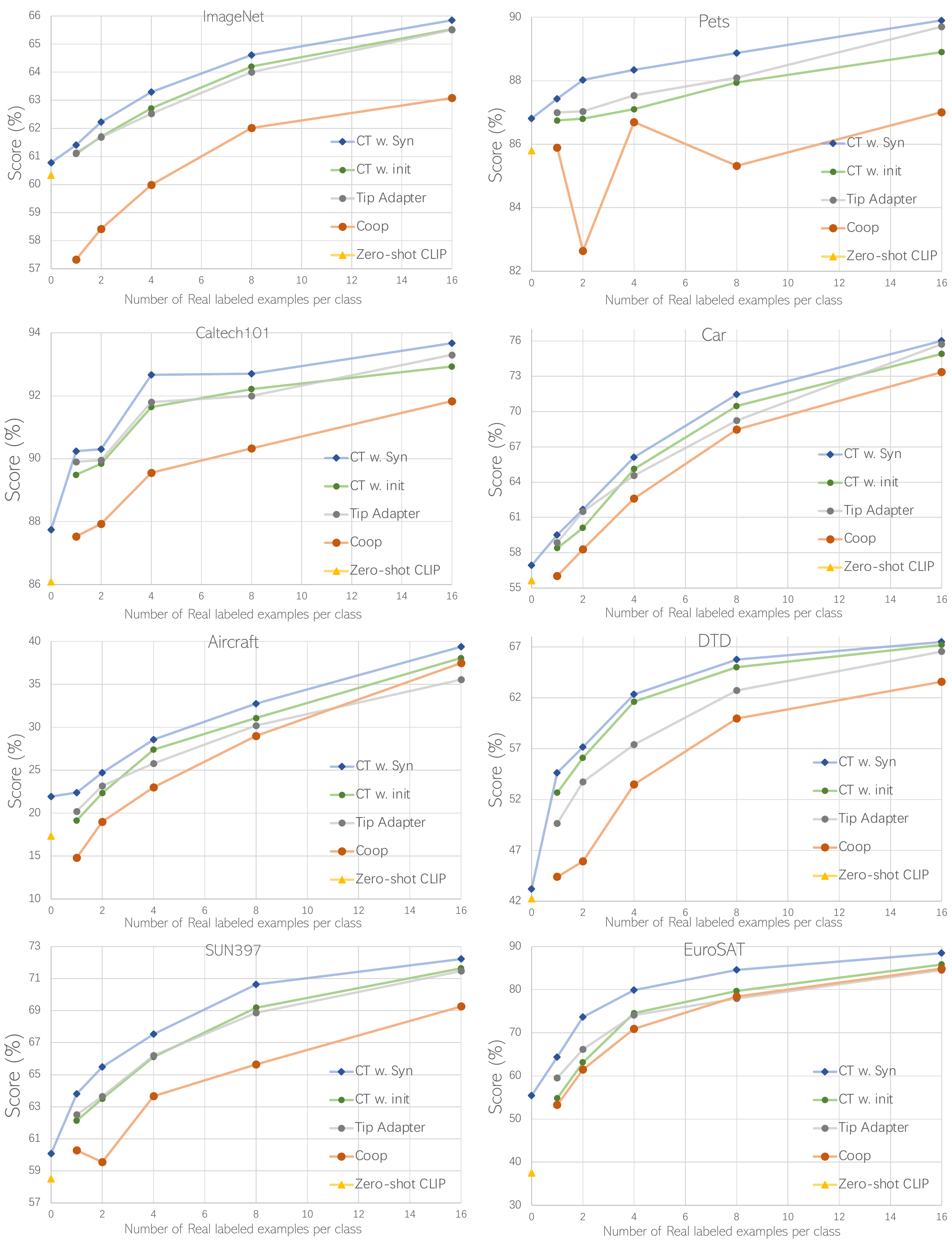} 
    \end{center}
 %    \vspace{-0.4cm}
       \caption{Main results for Few-shot tasks on all 8 datasets. }
    \label{fig: few-shot appendix} 
    \vspace{-0.6cm} 
    \end{figure}

Due to the limit of pages and space, we provide the results for few-shot tasks of all 8 datasets in Figure \ref{fig: few-shot appendix}. We observe similar results on various datasets that synthetic data could boost few-shot learning performance and achieve a new state-of-the-art performance. 

\subsection{real-world data with domain shift}
In order to understand why synthetic data may harm pre-trained image encoder, we experiment with real-world data with domain shifts. Specifically, we conduct experiments on pre-trained CLIP model and study two scenarios (Classifier tuning only, and end-to-end finetune) for the downstream ImageNet classification task. These experiments on conducted on three data sources: 1) in-domain real data: ImageNet data is used for training; 2) real-world data with domain shifts: ImageNet-Sketch data which has the same label space with ImageNet but exhibits a domain shift towards sketch are used for training; and 3) our synthetic data: synthetic data from GLIDE are used for training. The results are shown in the Table \ref{tab: domain shift real}. It can be clearly observed that real-world data with domain shifts behave similarly as synthetic data (i.e. end-to-end tuning cause slight harm to the pre-trained encoder) while the performance of in-domain data will benefit from end-to-end tuning. This suggests that domain gap is the main reason for harming the pre-trained image encoder.

We argue that synthetic data might have a better chance to overcome domain shifts in comparison with real-world data since we can customize and keep the label space of the synthetic data in line with the down-stream dataset. First, synthetic data can be made to tailor to a specific label space that the downstream task requires and reduce category shifts, which yet might be very challenging for real-world data. Second, a small amount of real-data can be leveraged to guide the data synthesis process to further alleviate domain shifts (namely Real Guidance in our few-shot settings), which also worth future in-depth explorations.

  \begin{table}[htbp] 
   \centering
    \vspace{-0.2cm}
   \begin{small}
       \setlength\tabcolsep{10pt} 
       \begin{tabular}{cccc}
           \bottomrule[1pt]
           Train data & ImageNet & ImageNet-Sketch  & Synthetic data \\ 
           \hline
           Classifier tuning &  70.09 & \highlight{60.50} & \highlight{60.78}  \\
           End-to-end finetune & \highlight{76.17} & 60.34 & 60.35 \\
           \toprule[0.8pt] 
       \end{tabular}
   \end{small}
   \vspace{-0.2cm}  
   \caption{Classifier tuning v.s. End-to-end finetune on different types of data. Zero-shot performance: 60.33. }
    \vspace{-0.2cm}
   \label{tab: domain shift real}
\end{table}

\subsection{Study of synthetic image number in zero/few-shot setting
}

Here, we provide the study for the number of synthetic images in the zero-shot and few-shot settings. 
Firstly, for zero-shot settings, we experiment with 1000/2000/4000 images per class on Cifar10 dataset. As shown in Table \ref{tab: syn num zsl}, we found 2000 to be a sufficient number while further increasing the number to 4000 only provide limited gains. 
Second, for few-shot settings, we study upon the settings of Eurosat dataset with 16 shot real images, and vary the number of synthesized images for each class from 400 to 1600. As shown in Table \ref{tab: syn num fsl},  we found 800 synthetic images for each class to be a good choice between performance and cost.

For both settings,  increasing the amount of training data further beyond a certain amount will not bring significant performance gain. The reasons can be attributed to the diversity and quality of data. We found that as the increase of the training data amount, the diversity of the data might not be scaled in a similar manner. Many redundant and similar samples will appear  with the increase of data amount. Effective approaches to increase data diversity and quality will help further improve model performance. 
For the few-shot setting, the performance reach a high value after using a small amount of synthetic data (400-shot). This is because the existence of real-data provide a strong guidance for training the classifier. And, the positive impacts of synthetic data are reduced where a small amount of synthetic data are sufficient to learn a good classifier.

  \begin{table}[htbp] 
   \centering
    \vspace{-0.2cm}
   \begin{small}
       \setlength\tabcolsep{10pt} 
       \begin{tabular}{ccccc}
           \bottomrule[1pt]
           syn shot & 0 & 1000& 2000& 4000 \\ 
           \hline
           Acc & 70.31 & 79.21 & \highlight{80.06} & \highlight{80.08} \\
           \toprule[0.8pt] 
       \end{tabular}
   \end{small}
   \vspace{-0.2cm}  
   \caption{Study of synthetic image number in the zero-shot setting on CIFAR-10 dataset. }
    \vspace{-0.2cm}
   \label{tab: syn num zsl}
\end{table}

  \begin{table}[htbp] 
   \centering
    \vspace{-0.2cm}
   \begin{small}
       \setlength\tabcolsep{10pt} 
       \begin{tabular}{ccccc}
           \bottomrule[1pt]
           syn shot & 0 & 400& 800 &1600\\ 
           \hline
           Acc & 85.83 & 88.28 & \highlight{88.47} &\highlight{88.49} \\
           \toprule[0.8pt] 
       \end{tabular}
   \end{small}
   \vspace{-0.2cm}  
   \caption{Study of synthetic image number in the few-shot setting on EuroSAT dataset with 16 shot real images. }
    \vspace{-0.2cm}
   \label{tab: syn num fsl}
\end{table}

\subsection{Claification of zero-shot classification performance difference
}

For all results, we use the official released CLIP model\footnote{https://github.com/openai/CLIP}. For our original paper, we conduct experiments using simple prompts ``a photo of a [CLASS]'' or ''a photo of a [CLASS], a type of [dataset type]'' (for fine-grained tasks) to better evaluate the performance gains from the data itself by excluding the influence of tuned prompt methods. However, in the original CLIP paper, they use specifically designed prompt ensembles for each dataset. This is the main reason that our baseline is lower than that of the original paper.

We also provide the results of using CLIP paper’s prompts for 13 datasets (only these 13 datasets out of 17 have reported results from CLIP’s paper) in Table \ref{tab: clip zsl compare}.
After using the same prompts as the CLIP's paper, the averaged performance on 13 datasets increased from 56.14\% to 56.33\%, closer to the 57.03\% of the CLIP's reported results.
There are still slight performance differences after using the same prompts, which we suspect to be a small reproduction problem of CLIP since we could match the reported zero-shot results from CoOp \citep{zhou2022learning} and Tip-adapter \citep{gao2021clip}.
We observe that on some datasets we achieve higher results than CLIP reported results (i.e., CIFAR-100, ImageNet, SUN397, Birdsnap, Flower, Pets) and on others we achieve lower results (i.e., CIFAR-10, Caltech101, Aircraft, Cars, Food, DTD, EuroSAT). The averaged zero-shot performance is similar (56.33\% v.s 57.03\%) and the performance boost from synthetic data of two different types of prompt is also similar (averaged performance boost on 13 datasets: 3.85\% v.s 4.17\%). We argue the slight performance differences do not affect the exploration of synthetic data.

  \begin{table}[htbp] 
   \centering
    \vspace{-0.2cm}
   \begin{small}
       \setlength\tabcolsep{5pt} 
       \begin{tabular}{cccccc}
           \bottomrule[1pt]
             & CLIP-RN50$^*$ &CLIP-RN50 & CLIP-RN50& CLIP-RN50$^\dagger$ & CLIP-RN50$^\dagger$+SYN  \\ 
           \hline
           CIFAR-10 & 75.6 & 70.31 & 80.06 \gain{9.75} & 71.59 & 80.23  \gain{8.64} \\
           CIFAR-100 & 41.6 &35.35 & 45.69  \gain{10.34} & 41.94 & 48.70 \gain{6.76} \\
           Caltech101 & 82.1 &86.09 & 87.74 \gain{1.65} & 79.99 & 82.34 \gain{2.35} \\
           ImageNet & 59.60 &60.33 & 60.78   \gain{0.45} & 60.33 & 60.78 \gain{0.45} \\
           SUN397 & 59.60 &58.51 & 60.07      \gain{1.56} &  60.23 & 60.47 \gain{0.24} \\
           Aircraft & 19.3 & 17.34 & 21.94   \gain{4.60}&17.07 & 21.78 \gain{4.71} \\
           Birdsnap & 32.6 &34.33 & 38.05   \gain{3.72} & 34.33& 38.05 \gain{3.72} \\
           Cars & 55.80 &55.63 & 56.93       \gain{1.30} & 55.70 & 57.65 \gain{1.95} \\
           Flower & 65.90 & 66.08 & 67.05     \gain{0.97} &66.08 & 67.05 \gain{0.97} \\
           Food & 81.10 &80.34 & 80.35       \gain{0.01} & 80.34 & 80.35 \gain{0.01} \\
           Pets & 85.40 &85.80 & 86.81       \gain{1.01} &  85.80 & 86.81 \gain{1.01} \\
           DTD & 41.7 &42.23 & 43.19           \gain{0.96} &  41.31 & 42.21 \gain{0.90} \\
           EuroSAT & 41.1 &37.51 & 55.37 \gain{17.86} & 37.52 & 55.35 \gain{17.83} \\
           \hline
           Average & 57.03 & 56.14& 60.31 \gain{4.17} & 56.33 & 60.10 \gain{3.85}\\
           \toprule[0.8pt] 
       \end{tabular}
   \end{small}
%   \vspace{-0.2cm}  
   \caption{CLIP-RN50$^*$: original CLIP paper results. CLIP-RN50: our results using simple prompt. CLIP-RN50$^\dagger$: our results using the same prompt ensemble as CLIP.}
    \vspace{-0.6cm}
   \label{tab: clip zsl compare}
\end{table}

\subsection{FID and classification accuracy with a pre-trained classifier for synthetic data measurement.
}
Frechet Inception Distance (FID) is a metric that calculates the distance between feature vectors from real and generated images, and lower scores have been shown to correlate well with higher-quality images. Here, we provide the FID scores and classification accuracy with a pre-trained CLIP ViT-B/16 model for measuring diversity and quality of synthesized images. 
We study with the Eurosat dataset. For the FID score, we do not have access to GLIDE training data, and thus we calculate FID between the downstream task ground-truth data and synthetic data generated by different strategies. As shown in Table \ref{tab: fid}, LE could largely reduce FID and provide large diversity while hurting class fidelity. After combing with CF, FID further reduces and we also achieve a higher class fidelity. For few shot settngs, RG could largely reduce FID and yield the best class fidelity.

  \begin{table}[htbp] 
   \centering
    \vspace{-0.2cm}
   \begin{small}
       \setlength\tabcolsep{10pt} 
       \begin{tabular}{cccccc}
           \bottomrule[1pt]
           syn data & z: B & z: LE & z: LE+CF & f: RF & f: RG\\ 
           \hline
           FID & 84.84 & 81.15 & 90.85 & 79.02 & \highlight{37.18} \\
           Acc & 44.27 & 35.84 & 48.03 & 46.95 & \highlight{48.76} \\
           \toprule[0.8pt] 
       \end{tabular}
   \end{small}
%   \vspace{-0.2cm}  
   \caption{FID and classification accuracy with CLIP-ViT-B/16. ``z'' for zero-shot and ``f'' for few-shot.}
    % \vspace{-0.2cm}
   \label{tab: fid}
\end{table}

\subsection{Examples of synthesized images from Language Enhancement strategy} \label{sec: LE A}
\begin{figure} 
    \begin{center}
    % \fbox{\rule{0pt}{2in} \rule{.9\linewidth}{0pt}} 
    \includegraphics[width=1.0\linewidth]{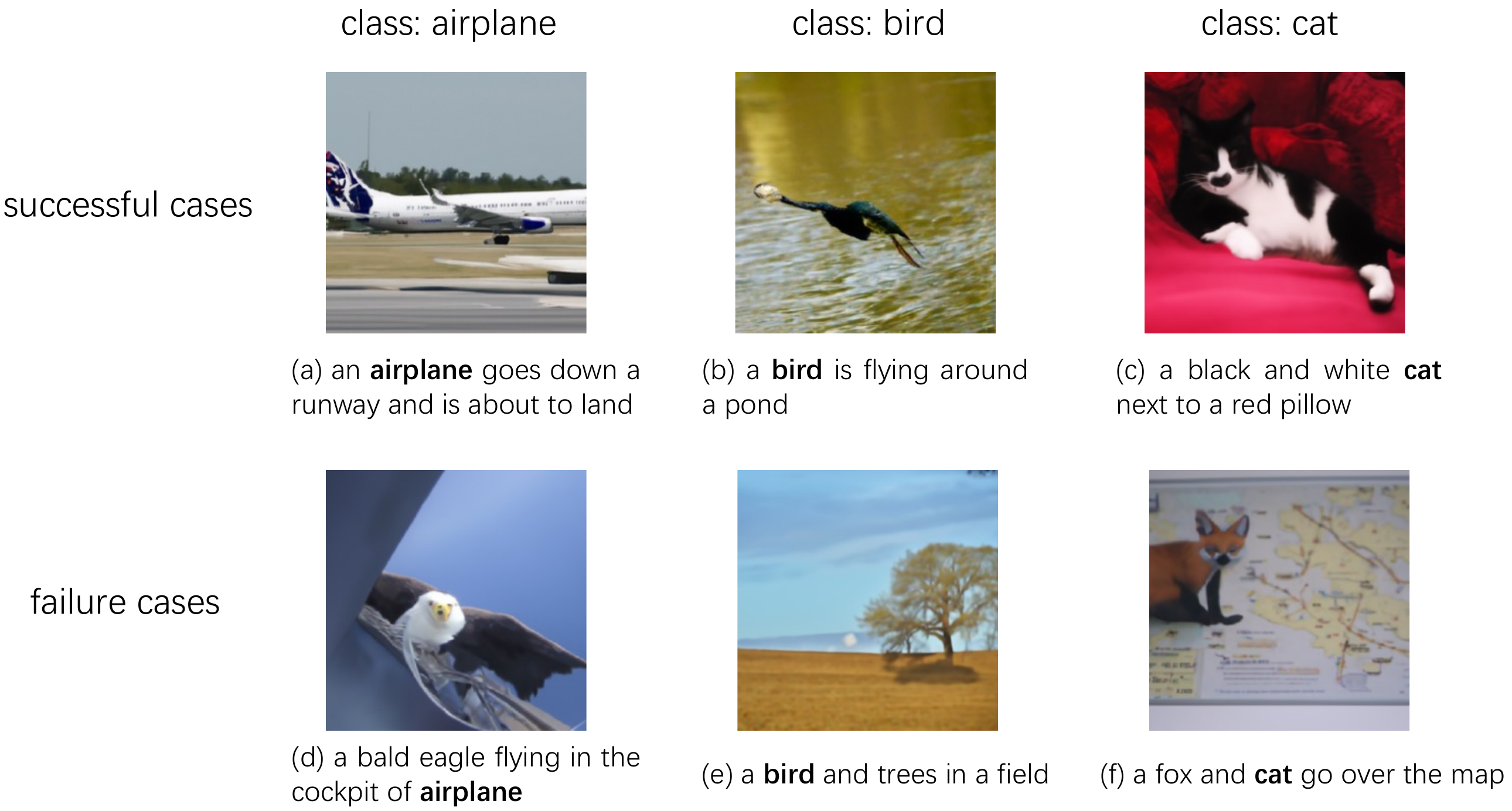} 
    \end{center}
 %    \vspace{-0.4cm}
       \caption{Examples of synthesized images from Language Enhancement strategy. }
    \label{fig: LE appendix} 
    % \vspace{-0.3cm} 
    \end{figure}

Here, we provide successful and failure examples of synthesized images from language enhancement strategy. As shown in Figure \ref{fig: LE appendix} (a) $\sim$ (c), language enhancement could introduce more diversity into the language prompts and lead to more diversified synthesized images for each class, such as introducing ``runway'' in (a), ``pond'' in (b), and ``red pillow'' in (c). However, we also observe failure cases after the language enhancement process. As we can see in Figure \ref{fig: LE appendix} (d) $\sim$ (f), after introducing some other items into the language prompts, the focus of the generated images may move to other objects rather than the target class. In some extreme failure cases we show here, the generated images may even not contain the desired class object.

\subsection{Visualization: synthetic data in zero-shot settings
}
\begin{figure}  
    \begin{center}
    % \fbox{\rule{0pt}{2in} \rule{.9\linewidth}{0pt}} 
    \includegraphics[width=1.08\linewidth]{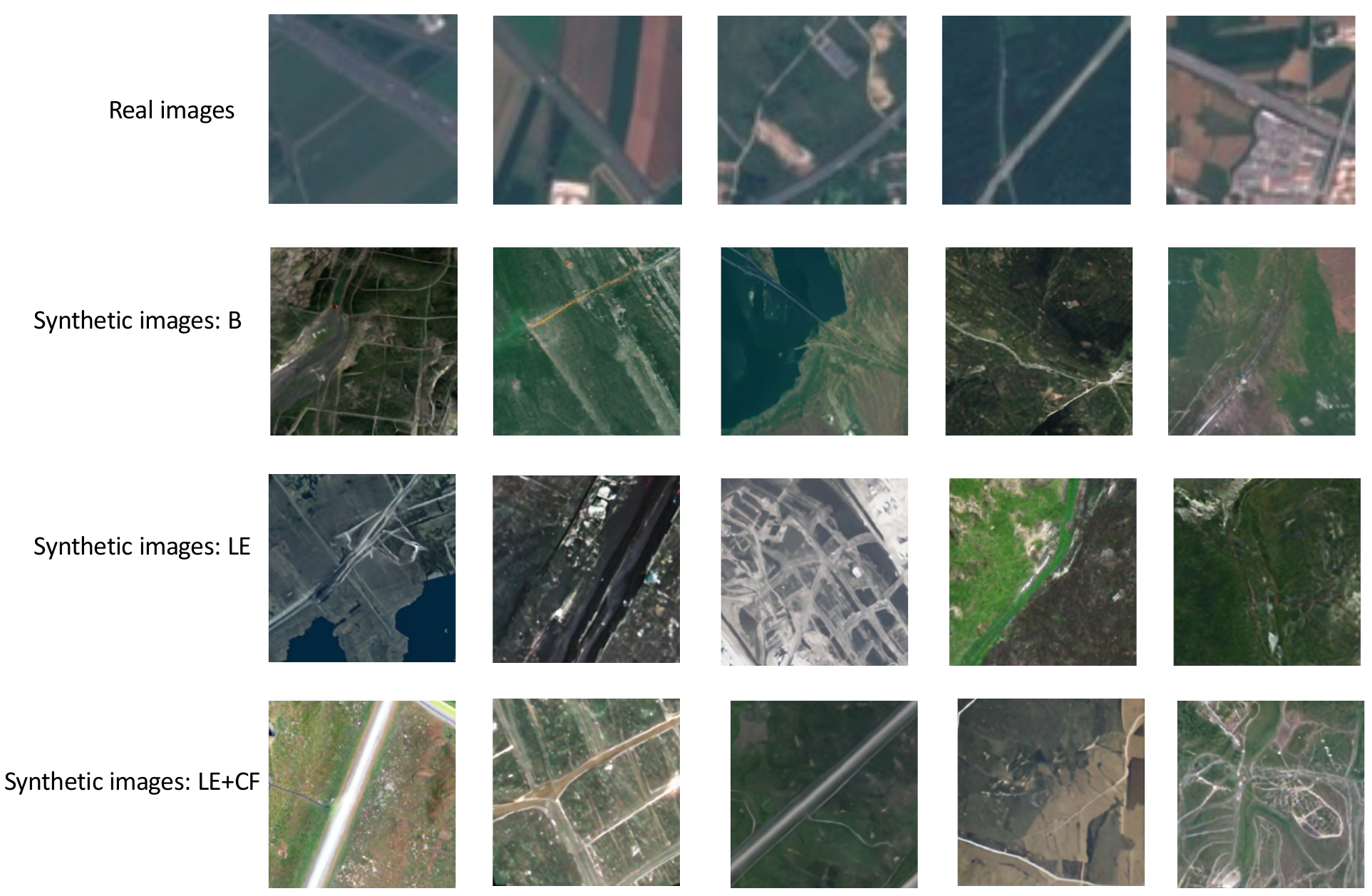} 
    \end{center}
 %    \vspace{-0.4cm}
       \caption{Visualization of different strategies of synthetic data in zero-shot settings. }
    \label{fig: vis_zsl} 
    % \vspace{-0.3cm} 
    \end{figure}

We provide the visual illustration of ground-truth real data and synthesized images by different strategies for the zero-shot settings, {\ie} basic (\textbf{B}), language enhancement (\textbf{LE}), and language enhancement and CLIP-based filtering (\textbf{LE+CF}). Here, we take the ``Highway or road'' class in EuroSAT dataset as an example. As shown in Figure \ref{fig: vis_zsl}, we could see that LE could help increase the diversity but may introduce noisy samples, but LE+CF could further select images with higher class fidelity and yields synthetic data with reduced domain gaps.

\subsection{Visualization: synthetic data in few-shot settings} \label{sec: RG A}

\begin{figure}  
    \begin{center}
    % \fbox{\rule{0pt}{2in} \rule{.9\linewidth}{0pt}} 
    \includegraphics[width=0.77\linewidth]{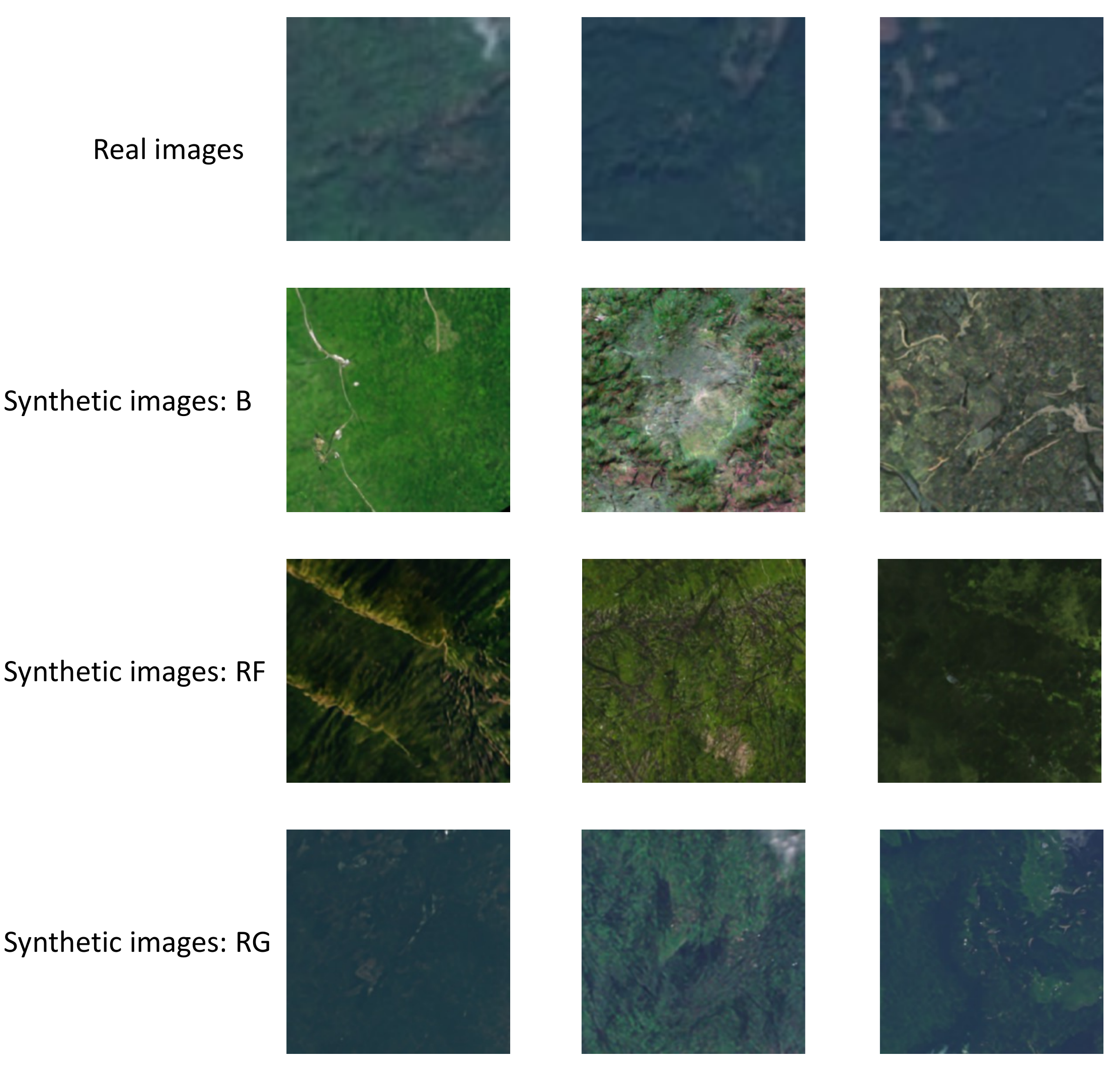} 
    \end{center}
 %    \vspace{-0.4cm}
       \caption{Visualization of different strategies of synthetic data in few-shot settings. }
    \label{fig: RG appendix} 
    % \vspace{-0.3cm} 
    \end{figure}

We provide the visual illustration of synthesized images by different strategies for the few-shot settings, {\ie} basic (\textbf{B}), real filtering (\textbf{RF}), and real guidance (\textbf{RG}) as well as real images of the same class for comparison. Here, we take the ``forest'' class in EuroSAT dataset as an example. As shown in Figure \ref{fig: RG appendix}, both \textbf{RF} and \textbf{RG} strategies produce images with reduced domain gap from the real images of the target domain. Further, \textbf{RG} significantly approaches the real images better than \textbf{RF}, demonstrating the effectiveness of the proposed \textbf{RG} method.  

\subsection{Visualization: synthetic data for different datasets
}
\begin{figure}  
    \begin{center}
    % \fbox{\rule{0pt}{2in} \rule{.9\linewidth}{0pt}} 
    \includegraphics[width=0.73\linewidth]{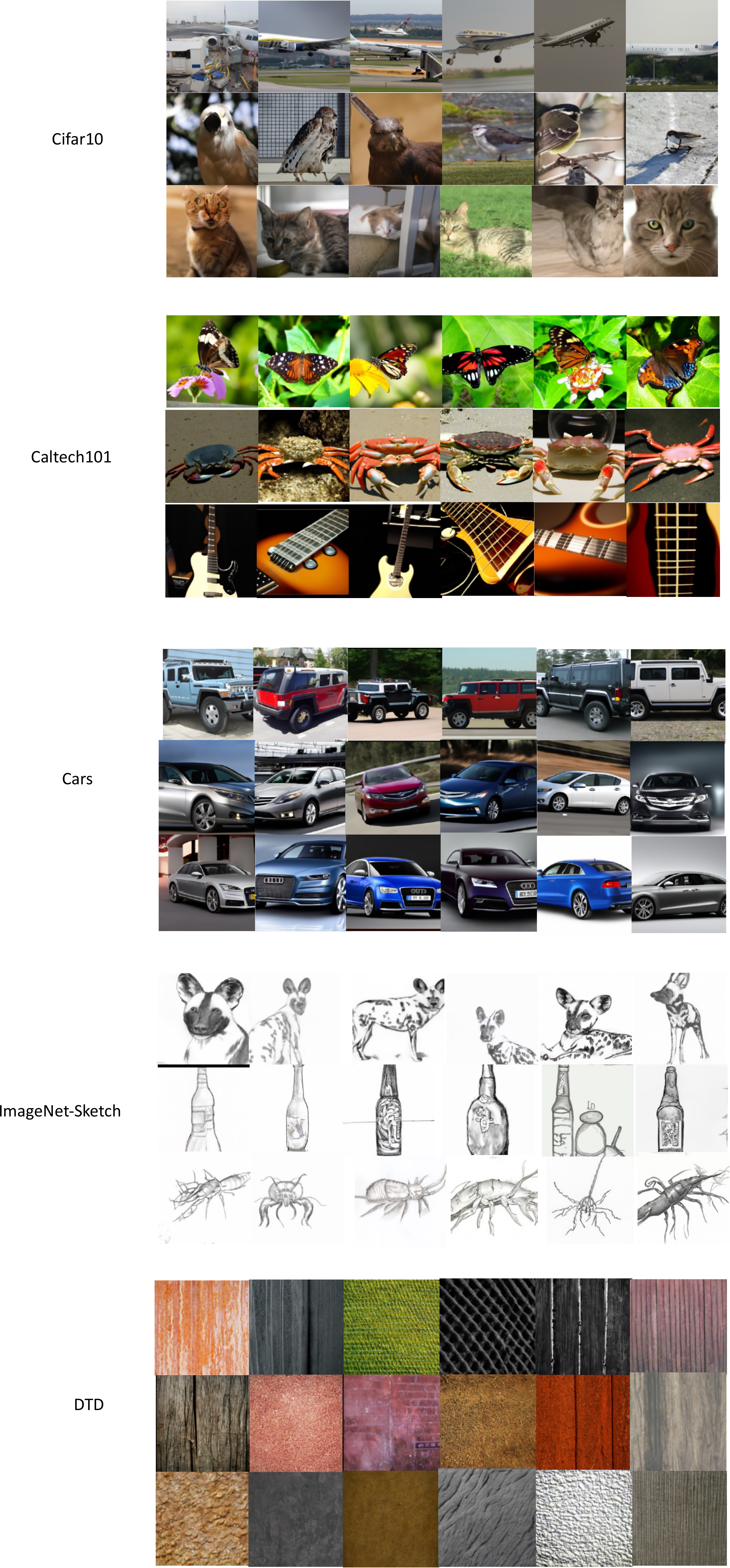} 
    \end{center}
 %    \vspace{-0.4cm}
       \caption{Visualization of different synthetic datasets. }
    \label{fig: syn datasets} 
    % \vspace{-0.3cm} 
    \end{figure}

Here, we provide synthesized images for different datasets (i.e., CIFAR10, Caltech101, Cars, ImageNet-Sketch, DTD) in Figure \ref{fig: syn datasets}. All images are randomly chosen rather than human-picked. Each row consists of images of the same class. We observe that for most datasets, synthesized images from the GLIDE model are of high quality, but there also exist cases that many unsatisfactory examples are generated, such as the DTD datasets.

We state that this is a limitation of the current text-image generation model that it may produce images of low quality for certain tasks, which mainly due to the domain gap between the training data of the generation model and the task. However, with the study of future text-image generation models, the quality of synthesized images is potentially growing higher constantly. Besides, the relatively lower quality images could be used for pre-training tasks for better improving performance of different down-stream tasks.

\section{Additional Details}

\algnewcommand{\algorithmicforeach}{\textbf{for each}}
\algdef{SE}[FOR]{ForEach}{EndForEach}[1]
  {\algorithmicforeach\ #1\ \algorithmicdo}% \ForEach{#1}
  {\algorithmicend\ \algorithmicforeach}% \EndForEach

  \renewcommand{\algorithmicrequire}{ \textbf{Input:}} 
  \renewcommand{\algorithmicensure}{ \textbf{Output:}}

% \subsection{Details of text-to-image diffusion model} 
% In this section, we provide a brief review and derivation of the basic denoising diffusion probabilistic model (\cite{ho2020denoising}) alongside with the text-to-image engine in GLIDE (\cite{nichol2021glide}).

\subsection{Denoising diffusion probabilistic model}
\label{ddpm}
Denoising diffusion probabilistic model (DDPM) learns the data distribution through introducing a series of latent variables and matching the joint distribution.
Formally, given a sample from the data distribution $x_0 \sim q(\mathbf{x}_0)$, a forward process $q\left(\mathbf{x}_{1: T} \mid \mathbf{x}_0\right)=\prod_{t=1}^T q\left(\mathbf{x}_t \mid \mathbf{x}_{t-1}\right)$ progressively perturbs the data with Gaussian kernels $q\left(\mathbf{x}_t \mid \mathbf{x}_{t-1}\right):=\mathcal{N}\left(\sqrt{1-\beta_t} \mathbf{x}_{t-1}, \beta_t \mathbf{I}\right)$, producing increasingly noisy latent variables $\mathbf{x}_{1}, \mathbf{x}_{2}, ..., \mathbf{x}_{T}$. Notably, $x_t$ can be directly sampled from $x_0$ thanks to the closed form:
\begin{equation}
q\left(\mathbf{x}_t \mid \mathbf{x}_0\right)=\mathcal{N}\left(\mathbf{x}_t ; \sqrt{\bar{\alpha}_t} \mathbf{x}_0,\left(1-\bar{\alpha}_t\right) \mathbf{I}\right)
\end{equation}
where $\alpha_t:=1-\beta_t$ and $\bar{\alpha}_t:=\prod_{s=1}^t \alpha_s$.
In general, the forward process variances $\beta_t$ are fixed and increased linearly from $\beta_1=10^{-4}$ to $\beta_T=0.02$. Besides, $T$ should be large ({\eg}, 1000) enough to ensure $q\left(\mathbf{x}_T \mid \mathbf{x}_0\right)\approx \mathcal{N}(0, \mathbf{I})$. 
Diffusion model aims to model the joint distribution $q\left(\mathbf{x}_{0:T}\right)$ which naturally involves a tractable sampling path for the marginal distribution $q\left(\mathbf{x}_{0}\right)$.

Specifically, the candidate distribution is formulated as a Markov chain with parameterized transition kernels:
\begin{equation}
    p_\theta\left(\mathbf{x}_{0: T}\right):=p\left(\mathbf{x}_T\right) \prod_{t=1}^T p_\theta\left(\mathbf{x}_{t-1} \mid \mathbf{x}_t\right),
    p_\theta\left(\mathbf{x}_{t-1} \mid \mathbf{x}_t\right):=\mathcal{N}\left(\mathbf{x}_{t-1} ; \boldsymbol{\mu}_\theta\left(\mathbf{x}_t, t\right), \boldsymbol{\Sigma}_\theta\left(\mathbf{x}_t, t\right)\right)
\end{equation}
The training is thus achieved by optimizing a variational bound of negative log likelihood: 
\begin{equation}
   \mathrm{E}_{q\left(\mathbf{x}_0\right)}\left[-\log p_\theta\left(\mathbf{x}_0\right)\right] \leq \mathrm{E}_{q\left(\mathbf{x}_{0: T}\right)}\left[-\log \frac{p_\theta\left(\mathbf{x}_{0: T}\right)}{q\left(\mathbf{x}_{1: T} \mid \mathbf{x}_0\right)}\right] =: L
\end{equation}
The loss term $L$ can be rewritten as:
\begin{equation}
\mathbb{E}_q[\underbrace{D_{\mathrm{KL}}\left(q\left(\mathbf{x}_T \mid \mathbf{x}_0\right) \| p\left(\mathbf{x}_T\right)\right)}_{L_T}+\sum_{t>1} \underbrace{D_{\mathrm{KL}}\left(q\left(\mathbf{x}_{t-1} \mid \mathbf{x}_t, \mathbf{x}_0\right) \| p_\theta\left(\mathbf{x}_{t-1} \mid \mathbf{x}_t\right)\right)}_{L_{t-1}} \underbrace{-\log p_\theta\left(\mathbf{x}_0 \mid \mathbf{x}_1\right)}_{L_0}]
\end{equation}

In practice, the core optimization terms are $L_{t-1} (t>1)$ that can be analytically calculated since both two terms compared in the KL divergence are Gaussians, {\ie},:
\begin{equation}
q\left(\mathbf{x}_{t-1} \mid \mathbf{x}_t, \mathbf{x}_0\right)=\mathcal{N}\left(\mathbf{x}_{t-1} ; \tilde{\boldsymbol{\mu}}_t\left(\mathbf{x}_t, \mathbf{x}_0\right), \tilde{\beta}_t \mathbf{I}\right),
p_\theta\left(\mathbf{x}_{t-1} \mid \mathbf{x}_t\right):=\mathcal{N}\left(\mathbf{x}_{t-1} ; \boldsymbol{\mu}_\theta\left(\mathbf{x}_t, t\right), \boldsymbol{\Sigma}_\theta\left(\mathbf{x}_t, t\right)\right)
\end{equation}
where $\tilde{\boldsymbol{\mu}}_t\left(\mathbf{x}_t, \mathbf{x}_0\right):=\frac{\sqrt{\bar{\alpha}_{t-1}} \beta_t}{1-\bar{\alpha}_t} \mathbf{x}_0+\frac{\sqrt{\alpha_t}\left(1-\bar{\alpha}_{t-1}\right)}{1-\bar{\alpha}_t} \mathbf{x}_t \quad$ and $\quad \tilde{\beta}_t:=\frac{1-\bar{\alpha}_{t-1}}{1-\bar{\alpha}_t} \beta_t$.
\cite{ho2020denoising} fix $\boldsymbol{\Sigma}_\theta\left(\mathbf{x}_t, t\right)=\sigma_t^2 \mathbf{I}$ during training, where $\sigma_t^2$ is set to be $\beta_t$ or $\tilde{\beta}_t$. Through reparameterization trick (\cite{kingma2013auto}) and empirical simplification (\cite{ho2020denoising}), the final training term is performed as follows:
\begin{equation}
L_{\text {simple }}(\theta):=\mathbb{E}_{t, \mathbf{x}_0, \epsilon}\left[\left\|\boldsymbol{\epsilon}-\boldsymbol{\epsilon}_\theta\left(\sqrt{\bar{\alpha}_t} \mathbf{x}_0+\sqrt{1-\bar{\alpha}_t} \boldsymbol{\epsilon}, t\right)\right\|^2\right]
\end{equation}
where $\epsilon\sim \mathcal{N}(0, \mathbf{I})$ and $t$ is uniformly sampled between $1$ and $T$.

After training, started from an initial noise map $x_T\sim p(\mathbf{x}_T)=\mathcal{N}(0, \mathbf{I})$, new images can be then generated via iteratively sampling from $p_\theta\left(\mathbf{x}_{t-1} \mid \mathbf{x}_t\right)$ using the following equation:
\begin{equation}
\mathbf{x}_{t-1}=\frac{1}{\sqrt{\alpha_t}}\left(\mathbf{x}_t-\frac{\beta_t}{\sqrt{1-\bar{\alpha}_t}} \boldsymbol{\epsilon}_\theta\left(\mathbf{x}_t, t\right)\right)+\sigma_t \mathbf{z}, \text { where } \mathbf{z} \sim \mathcal{N}(\mathbf{0}, \mathbf{I})
\end{equation}

\subsection{Text-to-image generation}
The text-to-image diffusion model extends the basic unconditional diffusion model by changing the target distribution $q(\mathbf{x_0})$ into a conditional one $q(\mathbf{x_0}\mid \mathbf{c})$, where $\mathbf{c}$ is a natural language description.
The derivation of the training terms and sampling procedure are similar to Sec. \ref{ddpm}, except that a conditioning signal $\mathbf{c}$ is included. Besides, following the improved DDPM (\cite{nichol2021improved}), $\boldsymbol{\Sigma}_\theta$ is also estimated in GLIDE (\cite{nichol2021glide}).

Especially, GLIDE employs a coarse-to-fine two-stage generation framework (\cite{nichol2021improved, saharia2022image}) with two guidance techniques for balancing mode coverage and sample fidelity, namely classifier guidance (\cite{dhariwal2021diffusion}) and classifier-free guidance (\cite{ho2022classifier}). 
Classifier guidance mainly relies on an extra trained noise CLIP model to provide feedback at intermediate sampling steps. Classifier-free guidance, on the other hand, randomly drops the text prompt with a fixed probability $p$ during the training, which can be viewed as a joint training of an unconditional model $\boldsymbol{\epsilon}_\theta\left(\mathbf{x}_t \mid \emptyset\right)$ ({\ie}, $\boldsymbol{\epsilon}_\theta\left(\mathbf{x}_t\right)$) and a conditional model $\boldsymbol{\epsilon}_\theta\left(\mathbf{x}_t \mid \mathbf{c}\right)$. At each sampling step, the model's output is actually performed using an extrapolation as follows:
\begin{equation}
\boldsymbol{\hat{\epsilon}}_\theta\left(\mathbf{x}_t \mid \mathbf{c}\right)=\boldsymbol{\epsilon}_\theta\left(\mathbf{x}_t \mid \emptyset\right)+s \cdot\left(\boldsymbol{\epsilon}_\theta\left(\mathbf{x}_t \mid \mathbf{c}\right)-\boldsymbol{\epsilon}_\theta\left(\mathbf{x}_t \mid \emptyset\right)\right)
\end{equation}
where $s$ is a guidance scale that can trade off sampling quality and diversity. In our work, we use classifier-free guidance with default setting $s=3$ for all experiments since it achieves better results than CLIP guidance. To speed up the sampling process, DDIM (\cite{song2020denoising}) is utilized which allows the model to produce high-quality images within few seconds. We follow the default settings in GLIDE and set $T=100$ in the coarse stage and $T=27$ in the upsampler stage.

\subsection{Real guidance (\textbf{RG}) strategy} \label{sec: RG}
We elaborate how we use few-shot in-domain real images to guide the generation process for few-shot settings. In a normal text-to-image generation process, a pure noisy image $x_{T}\sim \mathcal{N}(0, \mathbf{I})$ would be sampled first as the initialization of the reverse path. Then, the pretrained GLIDE model iteratively predicts a less noisy image $x_{t-1}$ $(t=T,T-1,...,1)$ using the given text prompt $c$ and the noisy latent image $x_t$ as inputs. In our case, we add noise to a reference image $x_0^{ref}$ such that the noise level corresponds to a certain time-step ${t_{\star}}$:
\begin{equation}
\label{eq:IGG}
x_{t_{\star}}^{ref}=\sqrt{\bar{\alpha}_{t_{\star}}} x_0^{ref}+\sqrt{1-\bar{\alpha}_{t_{\star}}} \mathbf{\epsilon}
\end{equation}
Then, rather than sampling from time-step $T$, we initialize the noisy latent variable as $x_{t_{\star}}^{ref}$ and begin our denoising process from time-step ${t_{\star}}$, as illustrated in Algorithm \ref{algo:IGG}. Note that the GLIDE model adopts a coarse-to-fine two-stage generation framework and involves classifier-free guidance. However, we omit them in Algorithm \ref{algo:IGG} for simplicity since our image-guidance strategy only modifies the start point and leaves the other settings unchanged. In this way, the generated images can share similar in-domain properties, and thus helping to close the domain gap. While small ${t_{\star}}$ could synthesis images which are more similar to the reference image, it results in low diversity, which harms the classifier's learning. In the case of a large ${t_{\star}}$, $x_{t_{\star}}^{ref}$ retains too little information from $x_0^{ref}$, causing the generated image to deviate from the domain. In our experiments, we conduct different trade-offs considering different few-shot settings. Empirically, we set ${t_{\star}}$ as 15, 20, 35, 40, and 50 for shot 16, 8, 4, 2, and 1, respectively.  

\begin{algorithm} 
    \footnotesize
    \caption{Real Guidance (\textbf{RG}) Strategy}
    \label{algo:IGG}
    \begin{algorithmic}[1]
        \Require 
        Reference image $x_0^{ref}$, text prompt $c$ and GLIDE model $(\mu_{\theta}, \Sigma_{\theta})$. 
        \Ensure
        Generated image $x_0$  

        \State \textcolor{gray}{ \# Noisy variable initialization}
        \State Select a time-step ${t_{\star}}\sim 1, 2, 3, ..., T$ and random noise $\epsilon \sim \mathcal{N}(0, \mathbf{I})$
        \State Obtain initial noisy image $x_{t_{\star}}:=x_{t_{\star}}^{ref}$ according to Eq. \ref{eq:IGG}
        \State \textcolor{gray}{ \# Random Sampling (could be replaced by DDIM for speed-up)  }
        \For {$s$ from ${t_{\star}}$ to $1$}
        \State $\mu, \Sigma \leftarrow \mu_\theta\left(x_s, s, c\right), \Sigma_\theta\left(x_s, s, c\right)$ 
        \State 
        $x_{s-1} \leftarrow$ sample from $\mathcal{N}\left(\mu, \Sigma\right)$
        \EndFor
        \State \Return $x_0$

    \end{algorithmic}
\end{algorithm}

\newpage
\subsection{soft-target cross-entropy loss} \label{sec: SCE A}

% /////////Settings for code//////////////
\lstset{language=Python}
\lstset{frame=lines}
\lstset{basicstyle=\ttfamily\footnotesize}
\lstset{escapeinside={<@}{@>}}
% ///////////////////////
Example code for soft-target cross-entropy loss is shown below. 

\begin{lstlisting}
def soft_target_cross_entropy(logits, target, labels, T=2):
    <@\texttt{\textcolor{gray}{\# T: temperature for soft targets.}}@>
    loss_func_CE = torch.nn.CrossEntropyLoss()
    CE = loss_func_CE(logits, labels)
    soft_targets = torch.softmax(target/T, dim=1)
    SCE = torch.sum(-soft_labels * F.log_softmax(x, dim=-1), dim=-1)
    <@\texttt{\textcolor{blue}{\textbf{loss = 0.5 * CE + 0.5 * SCE}}}@>
    return loss
    
\end{lstlisting}

\subsection{Implementation details} \label{sec: imple detail A}

\subsubsection{Zero-shot setting}
For text-to-image generation process, we adopt the default hyperparameters from the official GLIDE text-to-image code. The input text of the basic strategy is ``a photo of a [CLASS]'', and the input text of language enhancement strategy is ``a photo of a [SENTENCE]''. For language enhancement, we adopt an off-the-shelf word-to-sentence T5 model pre-trained on ``Colossal Clean Crawled Corpus'' dataset \citep{raffel2020exploring} and finetuned on CommonGen dataset \citep{lin2019commongen}. We generate 2000 synthetic images for each class in \textbf{B} and \textbf{LE}, and use a threshold of 1/N in \textbf{CF} where N is the number of classes. For \textbf{LE}, we generate 200 sentences for each class name.

For training on synthetic data for zero-shot recognition, we use AdamW \citep{loshchilov2017decoupled} optimizer and an initial learning rate of 0.002 that is decayed by the cosine annealing rule. We train for 30 epochs, and use weight decay of 0.1 and batch size of 512. For image preprocessing, we resize the image's short side to 224 while keeping the original aspect ratio.

For datasets in the zero-shot settings, we follow previous works \citep{zhou2022learning, gao2021clip} that use 11 datasets, and we excludes UCF101 since GLIDE exclude generating ‘person’ related content for privacy issues. Besides, we add another 7 popular datasets for more comprehensive evaluation. We do not conduct on all CLIP’s 27 datasets since our computing resources are limited, and we believe our 17 datasets are already enough to study the effectiveness of synthetic data for zero-shot settings.

\subsubsection{Few-shot setting}
For text-to-image generation process in few-shot settings, our basic strategy and Real filtering strategy both apply the same process as in the zero-shot settings; and for Real guidance strategy, the generation process is illustrated in Sec. \ref{sec: RG}. For synthetic image number, we generate 800 images per class for \textbf{RG} method to approximately match the number of images in \textbf{B} and \textbf{RF}.

For training methods in the few-shot settings, we provide the implementation details of phase-wise training and mix training. For phase-wise training, we utilize synthetic data and real data in two different phases, and the order of using synthetic data and real data also yields two variants. For syn$\rightarrow$real, we first tune the classifier on synthetic data for 30 epochs, and then tune on real data for 30 epochs; and change the order for real$\rightarrow$syn. For mix training, in each training iteration, we get a batch of real data input into the model and obtain the loss value of real part data, and also get a batch of synthetic data input into the model and get a loss value of synthetic part data, and then add two loss values as the final loss to do back propagation.

For training in the few-shot settings, we again use AdamW optimizer, weight decay of 0.1 and the cosine annealing rule. We use batch size of 32 for few-shot real images and 512 for synthetic images.
For phase-wise training, we train for 30 epochs for each stage and use an initial learning rate of 0.002. For mix training, we train for 30 epochs and use an initial learning rate of 0.001, where the loss value from real data and synthetic data are added with a 1:1 ratio in each iteration. For image preprocessing, we adopt the same strategy as zero-shot settings.

\subsubsection{Pre-training setting}
For pre-training settings, we adopt the \textbf{LE} strategy in zero-shot settings for generating massive amount of diversified synthetic pre-training data. For downstream-aware settings, we generate synthetic data by language prompts constructed from CIFAR-100 label space through the word-to-sentence model, and we generate synthetic data of {1.2M, 2.4M, 3.6M}. For downstream-agnostic settings, we generate from a generic label space of ImageNet-1K or ImageNet-2K, and data amount of {1.2M, 2.4M, 4M}.

For training details of downstream-aware synthetic pre-training on CIFAR-100 label space, we use AdamW optimizer, weight decay of 0.9, batch size of 512, training epochs of 90, and the cosine annealing rule for adjusting learning rate. For the initial learning rate, we use 1$e$-4 when pre-training from random initialization, and 1$e$-5 when pre-training from ImageNet pre-trained weights. For data augmentation, we adopt random cropping, resizing, and random horizontal flip. For transfering on CIFAR-100 dataset, we train for 200 epochs and use a SGD optimizer with an initial learning rate of 0.003, which is multiplied by 0.2 at 60, 120, 160 epochs. We use batch size of 128 and weight decay of 5$e$-4.   

For downstream-agnostic synthetic supervised pre-training with ResNet50 backbone, we train for 90 epochs and use a SGD optimizer with an initial learning rate of 0.2 for training from random initialization and 0.001 for training from ImageNet pre-trained weights. We multiply the initial learning rate by 0.1 every 30 epochs. We use batch size of 512 and weight decay of 1e-4. For data augmentation, we also adopt random cropping, resizing, and random horizontal flip. 

For downstream-agnostic synthetic supervised pre-training with DeiT-S backbone, we follow the training scripts for ImageNet pre-training and CIFAR-100 transfer learning in the official DeiT codebase\footnote{https://github.com/facebookresearch/deit} but only replace the pre-training data with our synthetic dataset. We do not change any hyper-parameters.

For downstream-agnostic synthetic self-supervised pre-training, {\ie} Moco v2, we follow the hyperparameters of the original implementation of the paper when training from random initialization. When initialized from ImageNet pre-trained weights, we use a small initial learning rate of 0.003 and keep other hyperparameters the same.

For transfer evaluation of object detection on PASCAL VOC 2012, we use the Faster R-CNN \citep{ren2015faster} detector and backbones are initialized by the pre-trained weights. All the setups follow the evaluation protocols in Moco \citep{he2020momentum}.

\end{document}